\documentclass[review]{elsarticle}
\usepackage{amssymb}
\usepackage{amsthm}
\usepackage{amsfonts}
\usepackage{graphicx}
\usepackage[utf8]{inputenc}
\usepackage[english]{babel}
\usepackage{algorithm,algorithmic,setspace}
\usepackage{csquotes}
\usepackage[varg]{newtxmath}
\usepackage{bm}
\usepackage{color}
\usepackage{multirow}
\usepackage{subcaption}
\usepackage{amsmath}
\usepackage{sansmath}
\usepackage{pifont}
\usepackage{url}
\usepackage{etoolbox}
\usepackage{lipsum}
\usepackage{boldline}
\usepackage{enumerate} 

\usepackage{lineno}
\modulolinenumbers[5]

\journal{Journal of \LaTeX\ Templates}
\usepackage{xspace}
\usepackage{soul}
\usepackage{enumerate}
\usepackage{enumitem}

\usepackage{hyperref}
\usepackage{cleveref}
\usepackage{amsthm}
\usepackage{color}

\def\Snospace~{\S{}}

\newcommand{\squishlist}{
	\begin{itemize}[noitemsep,nolistsep]
		\setlength{\itemsep}{-0pt}
	}
	\newcommand{\squishend}{
	\end{itemize}
}

\newcommand{\enumlist}{
	\begin{enumerate}[noitemsep,nolistsep]
		\setlength{\itemsep}{-0pt}
	}
	\newcommand{\enumend}{
	\end{enumerate}
}

\usepackage{xstring}
\newcommand{\PP}[1]{
	\vspace{2px}
	\noindent{\bf {#1}{.}}
}
\newcommand{\PC}[1]{
	\vspace{2px}
	\noindent{\bf \IfEndWith{#1}{:}{#1}{#1:}}
}
\newcommand{\IP}[1]{
	\vspace{2px}
	\noindent{\it \IfEndWith{#1}{.}{#1}{#1.}}
}
\newcommand{\IC}[1]{
	\vspace{2px}
	\noindent{\it \IfEndWith{#1}{:}{#1}{#1:}}
}

\newcommand{\boxbeg}{
	\noindent
	\vspace{0.3em}
	\begin{tabular}{|l|}
		\begin{minipage}{0.94\columnwidth}
			\vspace{0.1em}
			\noindent
		}
		
		\newcommand{\boxend}{
			\vspace{0.1em}
		\end{minipage}\\ 
	\end{tabular}
	\vspace{0.3em}
}

\newtheoremstyle{myDefinitionStyle}
{}
{}
{\normalfont}
{}
{\itshape}
{.}
{ }
{\thmname{#1}\thmnumber{ #2}\thmnote{ (#3)}}%

\newtheoremstyle{myTheoremStyle}
{}
{}
{\itshape}
{}
{\bfseries}
{.}
{ }
{\thmname{#1}\thmnumber{ #2}\thmnote{ (#3)}}%

\theoremstyle{myDefinitionStyle}

\theoremstyle{myTheoremStyle}

\usepackage{amsthm}

\begin{document}

\begin{frontmatter}

\title{Security and Privacy Enhanced Gait Authentication with Random Representation Learning and Digital Lockers}


\author[mymainaddress]{Lam Tran}


\author[thucaddress]{Thuc Nguyen}

\author[kimaddress]{Hyunil Kim}

\author[mymainaddress]{Deokjai Choi\corref{mycorrespondingauthor}}
\cortext[mycorrespondingauthor]{Corresponding author}
\ead{dchoi@jnu.ac.kr}

\address[mymainaddress]{Artificial Intelligence Convergence Department, Chonnam National University, South Korea}
\address[thucaddress]{Computer Science Department, Ho Chi Minh University of Sciences, Vietnam}
\address[kimaddress]{Robotics Engineering Department, Daegu Gyeongbuk Institute of Science \& Technology, South Korea}

\begin{abstract}
Gait data captured by inertial sensors have demonstrated promising results on user authentication.	
However, most existing approaches stored the enrolled gait pattern insecurely for matching with the validating pattern, thus, posed critical security and privacy issues.

In this study, we present a gait cryptosystem that generates from gait data the random key for user authentication, meanwhile, secures the gait pattern. 
\textit{First}, we propose a revocable and random binary string extraction method using a deep neural network followed by feature-wise binarization.
A novel loss function for network optimization is also designed, to tackle not only the intra-user stability but also the inter-user randomness.
\textit{Second}, we propose a new biometric key generation scheme, namely Irreversible Error Correct and Obfuscate (IECO), improved from the Error Correct and Obfuscate (ECO) scheme, to securely generate from the binary string the random and irreversible key.
The model was evaluated with two benchmark datasets as OU-ISIR and whuGAIT. 
We showed that our model could generate the key of $139$ bits from $5$-second data sequence with zero False Acceptance Rate (FAR) and False Rejection Rate (FRR) smaller than $5.441\%$.
In addition, the security and user privacy analyses showed that our model was secure against existing attacks on biometric template protection, and fulfilled irreversibility and unlinkability.
\end{abstract}


\end{frontmatter}

\linenumbers
\section{Introduction}\label{sec_intro}

Along with the evolution of microelectromechanical technology, inertial sensors (\textit{e.g.,} accelerometer, gyroscope) have been widely used and integrated into common mobile devices (\textit{e.g.,} smartphone, smartwatch).
This enables a practical, low-cost, and implicit mobile user authentication method, in which, the walking data captured by inertial sensors are exploited as the information sources to verify the user \cite{sprager2015inertial}.
The first inertial sensor-based gait recognition model was proposed by Ailisto in \cite{ailisto2005identifying}.
After that, many studies have been conducted and confirmed the promising of this approach \cite{sprager2015inertial, wan2018survey, sundararajan2019survey}.
However, existing models require storing the enrolled gait patterns in plaintext for matching with the verified data, thus pose critical security and privacy issues.

Biometric cryptosystems (BCSs) refer to the techniques that either generate a secret key from biometric data, or use biometric data to seal a random secret key \cite{rathgeb2011survey}.
In this approach, user authentication is carried out indirectly by matching the key that represent the encrypted biometric data.
Thus, the user privacy is also protected. 
Moreover, by using the key, BCS allows integrating biometric models to cryptographic schemes (\textit{e.g.,} symmetric encryption).
Several gait cryptosystems have been proposed (\textit{e.g.,} \cite{hoang2015gait, tran2017improving}) and demonstrated the potential of using BCS for gait, unfortunately, existed certain limitations. 
\textit{First}, they relied on Fuzzy Commitment Scheme (FCS) \cite{juels1999fuzzy} which has been specified as an unsecured solution when being deployed multiple times using the same biometric modality \cite{simoens2009privacy, kelkboom2010preventing,rathgeb2011statistical,scheirer2007cracking}.
\textit{Second}, unlinkability and irreversibility, specified as the mandatory requirements to ensure the user privacy \cite{information2011iso}, however, were not addressed in \cite{hoang2015gait, tran2017improving}.
\textit{Third}, the gait patterns were formed from the handcrafted features which had high intra-user variation and low inter-user discriminability, thus, existing models were inefficient and low authentication performance.

It is worth noting that, constructing a BCS model is such a challenging task, as it needs to resolve five difficulties as follows \cite{rathgeb2011survey,information2011iso}.
\textit{Correctness}: The enrolled user should be able to reproduce the generated key with high probability.
This is a challenging requirement, as the keys are required to match completely, however, biometric data always contain some variations. 
\textit{Secureness}: Only the enrolled user is able to reproduce his generated key. 
In addition, the keys generated from different users must be random from each other.
However, biometric data of different users usually show some common properties, and highly correlate to each other (\textit{i.e.}, not totally random as the key's expectation).
\textit{Revocability}: User should be able to change the key proactively, like changing a password.
This contradicts to biometric data which are persistent for a long time period.
\textit{Irreversibility}: It is computationally impractical to learn the biometric data from the keys and/or the public data (\textit{e.g.,} auxiliary data).
\textit{Unlinkability}: It is infeasible to identify with high certainty whether two given keys were constructed from the same or different users.

In this study, we propose a novel gait cryptosystem that securely and efficiently generates high-entropy keys from gait data, to be used for user authentication, and protects the user privacy.
\textit{First}, we observed that, raw gait data feature great intra-class variation but low inter-class discrimination which pose primary obstacles to achieve correctness and secureness. 
To address this, we propose a deep neural network that effectively extracts stable and discriminative features from raw gait data. 
The network is optimized using a new loss function which addresses not only the intra-class variation and inter-class discrimination of the extracted features, but also the randomness of the binary string derived from these features.
\textit{Second}, we present Irreversible Error Correct and Obfuscate (IECO), a BCS improved from Error Correct and Obfuscate (ECO) scheme \cite{canetti2016reusable}, to securely generate a random and irreversible key from the gait binary string.
The IECO scheme ensures the irreversibility which could not be achieved by ECO construction.
Moreover, we adopt random projection \cite{johnson1984extensions} as an extra protection layer, to provide revocability and unlinkability. 

In summary, our contributions in this study are:
\begin{itemize}
	\item 
	We proposed a method for extracting random binary string from gait segment using deep learning followed by feature-wise binarization. 
	The network was optimized by a novel loss function that tackles not only the intra-user stability and inter-user separability, but also the inter-user randomness (see \ref{ssec_dfe_block}). 
	Thus, the binary string extracted by our method is more random comparing to existing solutions (see \ref{sssec_loss_function_analyze}).
	\item 
	We introduced IECO, a new BCS that securely generates random and irreversible keys from biometric data (see \ref{ssec_revoca_bin_block}).
	Theoretical and experimental analyses on secureness and correctness of IECO were also conducted, to serve as an instruction for parameter fine-tuning (see \ref{sssec_param_analy}, \ref{sssec_symbol_size}).
	
	\item
	We conducted comprehensive experiments on the benchmark gait datasets (\textit{i.e.,} OU-ISIR \cite{ngo2014largest}, whuGAIT \cite{zou2020deep}). 
	The experimental results showed that our model can generate a key of 139 bits from $5$-second gait segment with zero FAR, and the FRR of $4.167\%$, $5.441\%$ for OU-ISIR, whuGAIT datasets, respectively (see \ref{ssec_ov_per}).
	We provided detailed security analyses for our scheme under practical attacks. 
	In addition, irreversibility and unlinkability evaluations were also conducted, and showed that our model could reliably protect the user privacy (see \ref{ssec_secu_pri_ana}).
	
\end{itemize}

Note that, the objective of this study is different from the gait-based pair-wise key generation researches which are well-summarized in \cite{bruesch2019security}.
In those studies, a nonce (\textit{i.e.}, key) is generated from gait data, to establish a secure communication channel between multiple devices wearing by a same person.
On the other hand, we generate a key, to be used for user authentication (before granting access to a service/resource). 
The objectives of two tasks are different, and each task has its own challenges and requirements to be addressed \cite{bruesch2019security}. 

\section{Related work}

This section briefly reviews the related researches on gait authentication and biometric cryptosystem. 

\subsection{Gait Authentication}

Gait recognition research was initiated with computer vision techniques 
\cite{niyogi1994analyzing, wang2003silhouette}. %
This approach offered a solution for video surveillance or access control in a specific area (\textit{e.g.,} building entrance) \cite{wan2018survey}.   
The evolution of sensing technology enabled a new gait recognition approach, which used the walking data obtained by Inertial Measurement Units (IMUs) attached on user's body \cite{ailisto2005identifying}.
This approach is promising for continuous user authentication on mobile devices \cite{wan2018survey}.
Many IMUs-based gait recognition studies have been conducted in the literature (\textit{e.g.,} \cite{sprager2015inertial, wan2018survey, tran2017improving, sprager2015efficient, gadaleta2018idnet, subramanian2019evaluation, tran2020data}). 
In the early stage, acceleration signals were used as main data source (\textit{e.g.,} \cite{ailisto2005identifying,gafurov2007gait,trivino2010application}).
The later researches additionally exploited the rotation rates obtained by gyroscope sensor to improve the performance (\textit{e.g.,} \cite{sun2014gait,gadaleta2018idnet}).
Moreover, some studies utilized multiple sensors placed in different positions of the user's body (\textit{e.g.,} \cite{giorgi2017try, dehzangi2017imu}).
Unfortunately, these studies did not provide any solution to protect the gait pattern, thus raised critical privacy and security issues \cite{rathgeb2011survey}.
Our preliminary works demonstrated the potential of using BCS for gait authentication \cite{hoang2015gait,tran2017improving}, however, existed several limitations on performance as well as security and privacy.

\subsection{Biometric Cryptosystem}

Several generic Biometric Cryptosystem (BCS) frameworks have been proposed, to allow associating noisy data with a secret key.
The first framework was proposed by Juels \textit{et al.} \cite{juels1999fuzzy}, named Fuzzy Commitment Scheme (FCS).
The main idea of FCS is to use biometric data to seal (\textit{i.e.,} bind) a random binary key, then adopt error correcting code (ECC) \cite{macwilliams1977theory} to handle the variation of biometric data.
After that, Dodis \textit{et al.} \cite{dodis2008fuzzy} introduced Fuzzy Extractor Scheme (FES), which was a generalization of FCS.
Another scheme proposed by Dodis was Fuzzy Vault Scheme (FVS) \cite{juels2006fuzzy}, that allowed sealing and revealing a secret key using sets of unordered biometric features.
Despite their effectiveness, these schemes exist certain vulnerabilities and suffer from several attacks when being deployed multiple times \cite{simoens2009privacy, kelkboom2010preventing,rathgeb2011statistical,scheirer2007cracking}.
In other words, these schemes could not ensure unlinkability and irreversibility  \cite{information2011iso}, thus could not remain secure when being deployed multiple times with the same biometric modality \cite{simoens2009privacy, blanton2013analysis}.

The later BCS models mainly focused on enabling reusability (\textit{a.k.a.,} revocability) (\textit{e.g.,} \cite{boyen2004reusable,apon2017efficient,canetti2016reusable}).
Boyen introduced a first reusable BCS framework, however, it relied on an impractical assumption that required no sensitive information was revealed from the exclusive OR (XOR) of the enrolled templates \cite{boyen2004reusable}. 
Recently, Canetti \textit{et al.} \cite{canetti2016reusable} introduced three fuzzy extractor schemes relying on Digital Lockers \cite{canetti2010symmetric}.
Among them, the first scheme, namely Sample Then Lock (STL) has been proved to be reusable under no assumption.
However, STL requires extremely large storage space for the helper data \cite{cheon2018reusable,zhu2019performance}.
The remaining two models, namely Error Correct and Obfuscate (ECO) and Condense then Fuzzy Extract (CFE), are not reusable.
Several studies attempted to reduce the helper data size in the STL model \cite{cheon2018reusable,zhu2019performance}.
In \cite{cheon2018reusable}, a threshold-based secret sharing scheme \cite{chen2016efficient} was adopted to increase the error handling capability of the STL model, thus, significantly improved the storage efficiency, unfortunately, also increased the computational cost.
The work in \cite{zhu2019performance} used ECC before adopting STL, to efficiently reduce the storage size.
However, this approach stored the offset between the biometric template and codeword for key reproduction.
This leads to several vulnerabilities as in FES and FCS.

Meanwhile, many BCS models on real biometric data have been proposed (\textit{e.g.}, 
fingerprint \cite{panchal2018novel}, face \cite{anees2018discriminative}, face and iris \cite{talreja2020deep}, gait \cite{hoang2015gait, tran2017improving}). 
Most initial studies focused on extracting from biometric data the deterministic and unique information, to be used as the key (\textit{e.g.}, \cite{hoang2015gait, tran2017improving, panchal2018novel}).
The later works started to address the revocability, unlinkability, and irreversibility (\textit{e.g.}, \cite{anees2018discriminative, talreja2020deep}). 
%


%

\section{Background}\label{sec_preliminaries}

\subsection{Digital Locker and Obfuscated Point Function}\label{ssec_dl_opf}

Digital Lockers (DL) are the symmetric encryption schemes that are computationally secure under multiple using times even with correlated and weak keys \cite{canetti2010symmetric}.
Let $s$ be a secret key and $v$ be a value. 
We mean $p=\mathsf{dLock}(s, v)$ as a DL algorithm that locks $v$ to $p$ using the key $s$, 
and $\mathsf{dUnlock}(s', p)$ as an unlock algorithm which returns $v$ if $s'=s$, otherwise $\bot$ with high probability.
With DL, obtaining any information of $v$ given $p$ is computationally difficult as guessing $s$.
In addition, a wrong $s'$ can be recognized with high probability.
We used the DL construction proposed in \cite{lynn2004positive}.
The lock function is $\mathsf{dLock}(s,v) = \{nonce, \mathsf{H}(nonce\| s) \oplus (v\|0^\gamma)\}$, where $\mathsf{H}(.)$ is a cryptographic hash function, $nonce$ is a nonce, and $\gamma$ is the security parameter.
Specifically, the algorithm can verify the correctness of $s'$ with the certainty of $1-2^{-\gamma}$.

Obfuscated Point Function (OPF) is a special DL, in which, the plaintext is empty (\textit{i.e.,} $v=\varnothing$).
We use $p=\mathsf{opLock}(s)$ as the OPF locking process.
Then, the unlock function $\mathsf{opUnlock}(s',p)$ will return $1$ if $s'=s$, and $0$ otherwise.

\subsection{Error Correcting Code}\label{subsec_BCH_code}

Error Correcting Codes (ECC) are the algorithms for constructing sequences of numbers in special ways so that any errors occurring after that (up to a certain number) can be corrected.
We use a family of ECC named BCH code \cite{macwilliams1977theory}, 
which allows correcting errors in any positions of the corrupted codeword. 
Let $z$ and $t$ be two positive integers satisfying $z \geq 3$ and  $t < 2^{z-1}$. 
There exists a BCH code, denoted as $\mathcal{C}(n,k,t)$, consisted of $2^k$ codewords $c$ of length $n$, where $n=2^z-1$, and $k$ satisfies $n-k \leq zt$.
Given any $m \in \{0,1\}^k$, we mean $c\leftarrow \mathsf{encode}_{\mathcal{C}}(m)$ as encoding $m$ to get $c\in \mathcal{C}$.
Let $c'\in\{0,1\}^n$ be an error version of $c$ such that $\mathsf{d}_H(c',c)\leq t$, where $\mathsf{d}_H(\cdot,\cdot)$ is the Hamming distance of the given strings.
Then, decoding $c'$ can correct all the errors to get $m$ (\textit{i.e.,} $m \leftarrow \mathsf{decode}_{\mathcal{C}}(c')$). 

\subsection{Error Correct and Obfuscate Scheme}\label{ssec_eco_scheme}
\begin{algorithm}[t]
	\setstretch{0.9}
	\caption{Key generation following ECO construction.}
	\label{alg_key_gen_ECO}
	\textbf{\textit{Input}}: the biometric template $\mathbf{s}$;\\
	\textbf{\textit{Output}}:
	the generated key $m$;
	the set of locked points $\mathcal{P}$;
	
	\begin{algorithmic}[1]
		\label{alg_ECO_generation}
		\STATE $m \overset{\$}\leftarrow \{0,1\}^{k};$\\
		\STATE $c \leftarrow \mathsf{encode}_{\mathcal{C}}{(m)};$\\

		\FOR {$i:=1$ to $n$ }
		\IF {$c_i == 1$}
		\STATE $p_i \leftarrow \mathsf{opLock}(s_i);$\\
		\ELSE
		\STATE $r_i \overset{\$}\leftarrow \mathcal{Z};$\\
		\STATE $p_i \leftarrow \mathsf{opLock}(r_i);$\\
		\ENDIF
		\STATE $\mathcal{P} \leftarrow \mathcal{P} \bigcup p_i;$
		\ENDFOR
		\RETURN $\{\mathcal{P}, m\};$
	\end{algorithmic}
\end{algorithm}
\begin{algorithm}[t]
	\setstretch{1.0}
	\caption{Key reproduction of the ECO construction.}
	\textbf{\textit{Input}}: the biometric template $\mathbf{s}'$;
	the set of locked points $\mathcal{P}$;
	
	\textbf{\textit{Output}}:
	the reproduced key $m'$;
	
	\begin{algorithmic}[1]
		\label{alg_ECO_reproduction}	
		\FOR {$i:=1$ to $n$ }
		\IF {$\mathsf{opUnlock}(s_i', p_i) == 1$}
		\STATE $c'_i \leftarrow 1;$\\
		\ELSE
		\STATE $c'_i \leftarrow 0;$\\
		\ENDIF
		\ENDFOR
		\STATE $m'\leftarrow \mathsf{decode}_{\mathcal{C}}(c');$\\
		\RETURN $m'$;
	\end{algorithmic}
\end{algorithm}
Error Correct and Obfuscate (ECO) is a fuzzy extractor scheme proposed by Canetti \textit{et al.} \cite{canetti2016reusable}, allows generating a random key from biometric template represented by a symbol string. 

Let $\mathcal{Z}$ be the symbol space (\textit{e.g.,} $\mathcal{Z}=\{0,1\}^\phi$, where $\phi$ is a small integer number).
Let a biometric template be represented as $\mathbf{s}=\begin{bmatrix}
	s_1 & \dots &s_i & \dots & s_n
\end{bmatrix},$
where $s_i \in \mathcal{Z}$.
The main idea of ECO is \textit{obfuscation} that locks either a biometric symbol $s_i$ or a random symbol $r_i\in\mathcal{Z}$ for each bit of the codeword (encoded from the key), according to the value of that bit. 
Without the key, it is unable to tell whether a given locked point is from a random symbol $r_i$ or a biometric symbol $s_i$.
The key generation process that outputs a key $m\in\{0,1\}^k$ and some helper data from $\mathbf{s}$ is summarized in algorithm \ref{alg_ECO_generation}.
Then, given another biometric template $\mathbf{s}'$, the key $m'$ could be reproduced as in algorithm \ref{alg_ECO_reproduction}.

ECO also utilizes ECC to handle the variation of biometric data, similar to FES.
However, in ECO, the distance between biometric template and the codeword is not stored as in FES.
Thus, the attacks on FES that exploit the helper data and ECC \cite{simoens2009privacy}, could not work on ECO model.
Despite the improvement, ECO does not offer reusability \cite{canetti2016reusable} (see \ref{sssec_security_improve}). 

\section{Gait Cryptosystem Framework}\label{sec_rkg_framework} 

\begin{figure*}[t]
	\centering
	\includegraphics[scale=0.029]{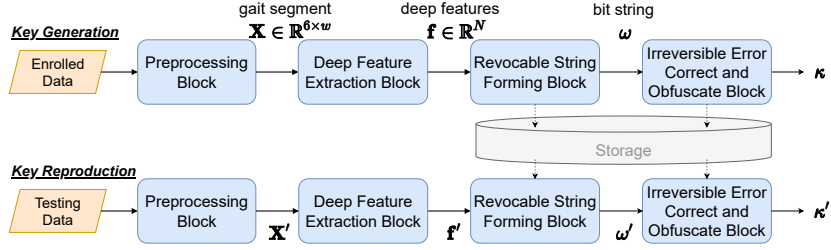}
	\caption{The processing blocks of our sensors-based gait cryptosystem.}
	\label{fig_model_archi}
\end{figure*}

\PP{Overview}
Figure \ref{fig_model_archi} sketches the overall model architecture, which operates in two phases, each phase is processed through four blocks.

\textit{Key generation (\textit{i.e.,} enrollment)}: First, the Preprocessing Block preprocesses and splits the raw data sequence into suitable segments.
Then, the Deep Feature Extraction (DFE) block extracts from the gait segment a representative feature vector $\mathbf{f}$ of length $N$.
Subsequently, the Revocable String Forming (RSF) block transforms $\mathbf{f}$ using random projection, then binaries the transformed template to a revocable bit string $\omega$ with the $\mathsf{sign}$ function. 
By random projection, multiple instances of $\omega$ could be produced, to enable key revocability.
Finally, the Irreversible Error Correct and Obfuscate (IECO) block generates a key $\kappa$ from $\omega$. 
Note that, in this phase, some helper data are extracted and publicly stored for key reproduction.
We show that it is impractical to learn neither the gait data nor the key from these data. 

\textit{Key reproduction (\textit{i.e.,} validation)}: The same processing blocks are performed on new gait data to obtain a key $\kappa'$, under the assistance of the helper data.
We now describe each processing block in detail.

\subsection{Data Preprocessing Block}
We use acceleration and gyroscope signals as the data source for key generation.
The data preprocessing and segmentation methods are referred from \cite{tran2021multi}, which are summarized as follows.

Let $\mathbf{A}=\begin{bmatrix}
\mathbf{a}_{t_1} & \mathbf{a}_{t_2} & \ldots & \mathbf{a}_{t_i} & \ldots 
\end{bmatrix}$ be a sequence of acceleration signals,
where $\mathbf{a}_{t_i}=\begin{bmatrix} a^X_{t_i} & a^Y_{t_i} & a^Z_{t_i}\end{bmatrix}^\top$ represents the acceleration forces acting along the $X$, $Y$, and $Z$ axes at the time $t_i$.
Similarly, we use $\mathbf{G}=\begin{bmatrix}
\mathbf{g}_{t'_1} & \mathbf{g}_{t'_2} & \ldots & \mathbf{g}_{t'_i} & \ldots 
\end{bmatrix}$ to represent a sequence of gyroscope signals,
where $\mathbf{g}_{t'_i}=\begin{bmatrix} g^X_{t'_i} & g^Y_{t'_i} & g^Z_{t'_i}\end{bmatrix}^\top$ represents the rotation rates around the $X$, $Y$, and $Z$ axes at the time $t'_i$.
The raw gyroscope signals are interpolated with respect to the acting time of acceleration signals \cite{schoenberg1973cardinal}, to overcome the asynchrony between the gyroscope and accelerometer.
Then, we combine the acceleration and gyroscope sequences to a six-channel data stream %
$
\begin{bmatrix}
\mathbf{x}_{t_1} & \mathbf{x}_{t_2} & \ldots & \mathbf{x}_{t_i} & \ldots 
\end{bmatrix},
$
where $\mathbf{x}_{t_i} = \begin{bmatrix}
a^X_{t_i} & a^Y_{t_i} & a^Z_{t_i} & g^X_{t_i} & g^Y_{t_i} & g^Z_{t_i}
\end{bmatrix}^\top$. 
The data stream is broken into fixed-length segments of $w$ signals, where $w$ is the window size, chosen so that each segment contains at least one gait cycle. 

\subsection{Deep Feature Extraction Block}\label{ssec_dfe_block}

Let $\mathbf{X}$ denote a gait data segment output by the Preprocessing Block.
In the DFE block, a deep network is used to extract a representative template $\mathbf{f}$ from $\mathbf{X}$.
This section first presents the network architecture, then describes the loss function for network optimization.%

\subsubsection{Network Architecture}\label{sssec_netarch}
\begin{figure}[t]
	\centering
	\includegraphics[scale=0.023]{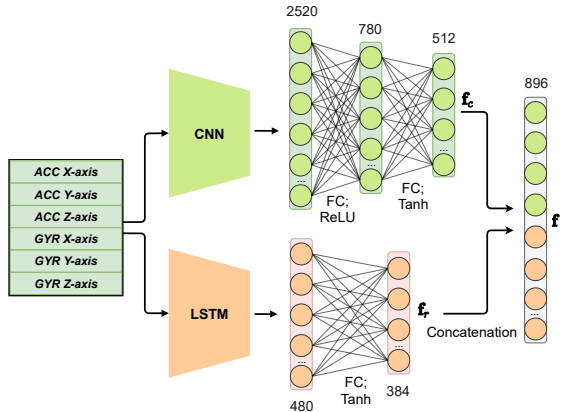}
	\caption{The deep network for extracting deep features from gait segment, where CNN is from \cite{delgado2018end} and LSTM is from \cite{tran2021multi}.}
	\label{fig_deep_model}
\end{figure}


The overall network architecture comprises of two branches as Convolutional Neural Network (CNN) and Long Short-term Memory (LSTM) network, extracting the features independently (see Figure \ref{fig_deep_model}).
\begin{itemize}
	\item 
	\textit{CNN}: This branch extracts a spatial feature template $\mathbf{f}_c$ from the gait segment $\mathbf{X}$.
	We leverage the CNN architecture proposed in \cite{delgado2018end} and make appropriate modifications for extracting representative template.
	Specifically, from the original network, we discard the last softmax layer, then add two fully-connected (FC) layers.
	The first FC layer scales the feature dimension from $2520$ to $780$, and uses rectified linear unit (ReLU) as the activation function.
	The second FC layer computes the representative template $\mathbf{f}_c$ of size $512$ from the output of the first FC layer.
	In this layer, $\mathsf{tanh}$ is used as the activation function to scale the feature value to the range $(-1,1)$. 
	\item 
	\textit{LSTM}: We utilize the LSTM network proposed in \cite{tran2021multi} for extracting from $\mathbf{X}$ the temporal features $\mathbf{f}_r$.
	We also discard the last softmax layer of the original network,
	then, add a new FC layer, to map the extracted features to a template $\mathbf{f}_r$ of size $384$.
	$\mathsf{Tanh}$ also is used as the activation function.
\end{itemize}
The representative template $\mathbf{f}$ is formed by concatenating $\mathbf{f}_r$ and $\mathbf{f}_c$, $\mathbf{f} = \mathbf{f}_c \| \mathbf{f}_r$.

\subsubsection{Random Representation Learning}\label{sssec_trai_optim}

Each branch of the network is trained separately, using the same optimization method.
Here, to simplify the notation, we use $\mathbf{f}$ as the output of CNN or LSTM network (\textit{i.e.,} $\mathbf{f}=\mathbf{f}_c$, or $\mathbf{f}=\mathbf{f}_r$) when explaining the optimization method.

Several optimization functions have been proposed to extract representation string for image retrieval (\textit{e.g.,} \cite{chen2019deep,ma2021learning}).
However, these studies addressed the task of semantic-preserving representation learning, in which, the Hamming distance of the extracted strings should reflect the similarity of the information sources.
Such methods are inappropriate to be used in a BCS model. 
A secure BCS model expects that, \textit{(I)} the binary strings extracted from the same user should be close to each other, and \textit{(II)} those from different users should randomly differ $50\%$ (\textit{i.e.,} differ from each other $50\%$, and the different positions are random and unpredictable). 
The requirement \textit{(II)} is critically important, and strongly affects the model's security.

We introduce a new optimization function for learning the representative templates $\mathbf{f}$ so that the binary string $\omega$ obtained from $\mathbf{f}$ satisfies two requirements above.
First, we consider the triplet loss \cite{schroff2015facenet} which optimizes the feature templates to be close when extracted from same user, and far for different users.
Let ${\mathbf{f}}_{u}$ and $\mathbf{f}'_u$ be two feature templates extracted from a user $u$, and $\mathbf{f}_v$ be the template of another user $v \neq u$. 
Then, the triplet loss is computed as:
\begin{equation}\label{eq_triplet}
L_T=\max(\mathsf{d}_E(\mathbf{f}_u, \mathbf{f}'_u)-\mathsf{d}_E(\mathbf{f}_u, \mathbf{f}_v)+ \delta,0),
\end{equation}
where $\mathsf{d}_E(\cdot,\cdot)$ means the Euclidean distance, and $\delta$ is the desired margin.
By minimizing $L_T$, $\mathsf{d}_E(\mathbf{f}_u, \mathbf{f}'_u)$ is pushed to zero, and $\mathsf{d}_E(\mathbf{f}_u, \mathbf{f}_v)$ tends to be larger than $\mathsf{d}_E(\mathbf{f}_u, \mathbf{f}'_u)+ \delta$.
The use of triplet loss solves the requirement \textit{(I)} and pushes the templates of different users as far as the margin $\delta$, however, could not ensure the requirement \textit{(II)}.
To address this, we propose additional criterion as follows.
As $\mathsf{tanh}$ is used as the activation function in the last layer (see \ref{sssec_netarch}), the value of each feature is scaled to the range $(-1,1)$.
Let $f_{u,i}$ and $f_{v,i}$ be the feature $i$ of $\mathbf{f}_u$ and $\mathbf{f}_v$, respectively.
As $\mathsf{sign}$ is used as the binarization function (see \ref{sssec_binary}), two users $u$ and $v$ will have a same bit $i$ if ${f}_{u,i}{f}_{v,i} > 0$, and different bit if ${f}_{u,i}{f}_{v,i} < 0$.
Thus, to achieve \textit{(II)}, the number of positions having ${f}_{u,i}{f}_{v,i} > 0$ should be approximate to the number of positions having ${f}_{u,i}{f}_{v,i} < 0$.
This can be performed by minimizing
\begin{equation}\label{eq_random_loss}
L_R= |\sum_{i=1}^N {f}_{u,i}{f}_{v,i}|=|\langle \mathbf{f}_u,\mathbf{f}_v\rangle|,
\end{equation}
where $\langle \cdot ,\ \cdot \rangle$ is the inner product.
Combining \eqref{eq_triplet} and \eqref{eq_random_loss}, we have 
\begin{equation}\label{eq_final_loss}
L = \max(\alpha \mathsf{d}_E(\mathbf{f}_u, \mathbf{f}'_u)-\mathsf{d}_E(\mathbf{f}_u, \mathbf{f}_v)+ \beta |\langle \mathbf{f}_u,\mathbf{f}_v\rangle| + \delta, 0),
\end{equation}
where $\alpha$ and $\beta$ are the regularization parameters. 
By minimizing $L$, the feature templates of the same user will get closer (due to the constraint of $\mathsf{d}_E(\mathbf{f}_u, \mathbf{f}'_u)$), thus the intra-class variation of the binary string $\omega$ is also reduced.
Meanwhile, the feature templates of different users will be more different in a way that the number of positions having ${f}_{u,i}{f}_{v,i} > 0$ is approximate to the number of cases having ${f}_{u,i}{f}_{v,i} < 0$ (due to the constraint $-\mathsf{d}_E(\mathbf{f}_u, \mathbf{f}_v)+ |\langle \mathbf{f}_u,\mathbf{f}_v\rangle|$).

\subsection{Revocable String Forming Block}\label{ssec_revoca_bin_block}
The Revocable String Forming (RSF) block transforms $\mathbf{f}$ to a revocable binary string $\omega \in \{0, 1\}^{\phi n}$, where $\phi$ is the symbol size, and $n$ is the codeword length (see \ref{ssec_eco_block}).
This task is performed in two steps as follows.

\subsubsection{Random Projection}
First, instead of using $\mathbf{f}$ to generate the key, we adopt Random Projection (RP) to provide revocability.
RP is a dimensional reduction technique inspired from the Johnson-Lindenstrauss Lemma \cite{johnson1984extensions}, which allows projecting vector templates into a random sub-space while approximately preserving their pair-wise similarity.
Specifically, given a feature template $\mathbf{f} \in \mathbb{R}^N$, we project it to a $K$-dimensional vector 
$\hat{\mathbf{f}}$ as
$\hat{\mathbf{f}} = \mathbf{f} \times \mathbf{R},$
where $K=N-1$, and $\mathbf{R}$ is an $N\times K$ matrix whose entries are independent and identically distributed according to a Gaussian distribution of zero mean and $\frac{1}{N}$ variance.
After this step, $\mathbf{\hat{f}}$ is used to generate the key while $\mathbf{f}$ is discarded.

\subsubsection{Binarization and Reliable Bits Selection}\label{sssec_binary}
To increase the stability, $M$ projected templates
(derived from $M$ segments projected with the same RP matrix)
are used to construct one binary string.
For each feature $i$ $(1\leq i \leq K)$, we compute the mean $\bar{{f}}_i= \frac{1}{M}\sum_{j=1}^M {\hat{f}}^j_{i}$, where $\hat{{f}}^j_{i}$ is the feature $i$ of the projected template $\hat{\mathbf{f}}^j$ ($1\leq j \leq M$).
Then, $\omega_i$ is determined:
\begin{equation}
\omega_i = \mathsf{sign}(\bar{f}_i)=\begin{cases}
1 & \text{if}\  \bar{f}_i \geq 0,\\
0 & \text{otherwise}.
\end{cases}    
\end{equation}
Subsequently, we statistically select $\phi n$ reliable bits which have low intra-class error.
The reliability of a feature is computed by
$r_i = -\frac{1}{M-1}\sum_{j=1}^M (\hat{f}^j_i-\bar{f}_i)^2$.
Then, $\omega$ is formed from $\phi n$ bits having highest reliability.

\subsection{Irreversible Error Correct and Obfuscate Block}\label{ssec_eco_block}
We present here the IECO scheme, which generates from $\omega$ an irreversible key $\kappa$. 
The overall idea of IECO is to use a nonce $m$ to lock the key $\kappa$ by a DL, then use the reliable string $\omega$ to seal $m$ following the ECO construction. 

\subsubsection{Key Generation}
Let $\phi$ be the symbol size, \textit{i.e.,} $\mathcal{Z}=\{0,1\}^\phi$.
Given $\omega$, a symbol string $\mathbf{s} \in \mathcal{Z}^n$ is constructed to be used as the input of the IECO scheme, where each symbol $s_i$ is formed from $\phi$ consecutive bits of $\omega$,
$s_i = \omega_{(i-1)\phi+1} \| \ldots \| \omega_{i\phi}.$
Then, a key $\kappa$ is generated from $\mathbf{s}$ as follows:

\begin{enumerate}
	\item[\textit{(i)}] 
	First, $\kappa\in\{0,1\}^k$ is generated randomly.
	\item[\textit{(ii)}]
	Next, a nonce $m\in \{0,1\}^k$ is encoded to a codeword $c$, 
	$c\leftarrow \mathsf{encode}_{\mathcal{C}}(m)$.
	Meanwhile, $m$ is used to lock the key $\kappa$ to $L_{\kappa}$ using a DL, $L_{\kappa}\leftarrow \mathsf{dLock}(m,\kappa)$.
	%
	%
	\item [\textit{(iii)}] For each bit $c_i$ ($1\leq i\leq n$), a locked point $p_i$ is created 
	\begin{equation}
	p_i = 
	\begin{cases}
		\mathsf{opLock}(s_i) & \text{if}\  c_i=1,\\
		\mathsf{opLock}(r_i) & \text{otherwise},
	\end{cases}    
	\end{equation}
	%
	where $r_i$ is generated randomly from $\mathcal{Z}$ so that $r_i \neq s_i$.		
%
%
	We store $L_\kappa$ and $p_i$ for key reproduction, while $\mathbf{s}$, $m$, $\kappa$ and $r_i$ are discarded. 
\end{enumerate}

\begin{figure*}[t]
	\centering
	\begin{subfigure}{.5\textwidth}
		\centering
		\includegraphics[scale=0.0221]{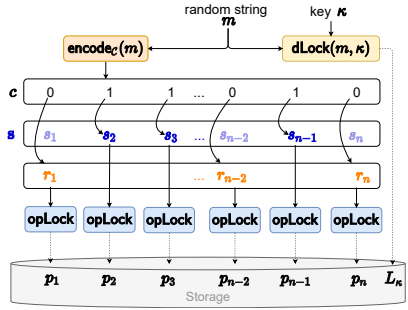}
		\caption{Key generation.}
		\label{fig_key_gen}
	\end{subfigure}%
	\begin{subfigure}{.5\textwidth}
		\centering
		\includegraphics[scale=0.0221]{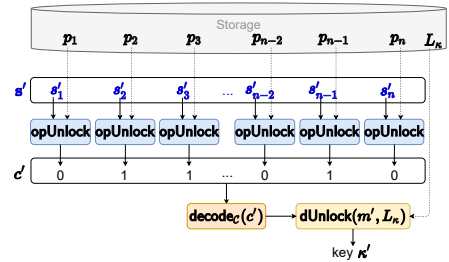}
		\caption{Key reproduction.}
		\label{fig_key_rep}
	\end{subfigure}%
	\caption{An illustration of key generation and reproduction following the IECO construction.}
	\label{fig_key_gen_rep}
\end{figure*}

We provide the pseudo-code for the key generation process in Algorithm \ref{alg_eco_gen_mor}, and depict an example in Figure \ref{fig_key_gen}.

\begin{algorithm}[t]
	\setstretch{0.9}
	\caption{Key generation using IECO scheme.}
	\textbf{\textit{Input}}: 
		the symbol string $\mathbf{s}$;\\
	\textbf{\textit{Output}}:
		the generated key $\kappa$; the locked points $\mathcal{P}$. 
		
	\begin{algorithmic}[1]
		\label{alg_eco_gen_mor}
		\STATE $\kappa \overset{\$}\leftarrow \{0,1\}^{k};$\\
		\STATE $m \overset{\$}\leftarrow \{0,1\}^{k};$\\
		\STATE $L_\kappa \leftarrow \mathsf{dLock}(m, \kappa);$\\
		
		\STATE $c \leftarrow \mathsf{encode}_\mathcal{C}{(m)};$\\
		\FOR {$i:=1$ to $n$ }
		\IF {$c_i == 1$}
		\STATE $p_i \leftarrow \mathsf{opLock}(s_i);$\\
		\ELSE
		\STATE $r_i \overset{\$} \leftarrow \{z\in \mathcal{Z}|z\neq s_i\};$\\
		\STATE $p_i \leftarrow \mathsf{opLock}(r_i);$\\
		\ENDIF
		\STATE $\mathcal{P} \leftarrow \mathcal{P} \bigcup p_i;$\\
		\ENDFOR
		\RETURN $\{\mathcal{P}, L_\kappa, \kappa\};$
	\end{algorithmic}
\end{algorithm}


\subsubsection{Key Reproduction}
Given a validating binary string $\omega'$, a symbol string $\mathbf{s}'$ is constructed.
Then, $\mathbf{s}'$ is used to reproduce the generated key as follows: 
\begin{enumerate}
	\item[\textit{(i)}]
	First, each bit of the codeword $c'$ is determined
	\begin{equation}
	c'_i = 
	\begin{cases}
	1 & \text{if} \ \mathsf{opUnlock}(s'_i, p_i)=1 ,\\
	0 & \text{otherwise}.
	\end{cases} 
	\end{equation}
	
	\item[\textit{(ii)}]
	Then, a nonce $m'$ is obtained from $c'$ by $m' \leftarrow \mathsf{decode}_{\mathcal{C}}(c')$.
	\item[\textit{(iii)}]
	Finally, $\kappa'$ is retrieved as $\kappa'\leftarrow \mathsf{dUnlock}(m',L_\kappa)$.
	In this step, $\kappa'$ is identical to $\kappa$ if $m'=m$, otherwise $\kappa'=\varnothing$ with the certainty of $1-2^{-\gamma}$.
\end{enumerate} 
Algorithm \ref{alg_eco_rep_mor} summarizes the pseudo-code for key reproduction, and Figure \ref{fig_key_rep} sketches an example of it.

\begin{algorithm}[t]
	\setstretch{0.9}
	\caption{Key reproduction with the IECO scheme.}
	\textbf{\textit{Input}}: 
		the biometric template $\mathbf{s}'$;
		the set of locked points $\mathcal{P}, L_{\kappa}$.\\
	\textbf{\textit{Output}}:
		the reproduced key $\kappa'$.
	\begin{algorithmic}[1]
		\label{alg_eco_rep_mor}
		\FOR {$i:=1$ to $n$ }
		\IF {$\mathsf{opUnlock}(s_i',p_i) == 1$}
		\STATE $c'_i \leftarrow 1;$\\
		\ELSE
		\STATE $c'_i \leftarrow 0;$\\
		\ENDIF
		\ENDFOR
		\STATE $m'\leftarrow \mathsf{decode}_{\mathcal{C}}(c');$\\
		\STATE $\kappa' \leftarrow \mathsf{dUnlock}(m', L_\kappa);$
		\RETURN $\kappa';$
	\end{algorithmic}
\end{algorithm}

\subsubsection{Correctness and Secureness Analysis}\label{sssec_param_analy}

In this section, we analyze the impact of IECO's parameters to the correctness and secureness, to provide instructions for parameter fine-tuning.
\textbf{Correctness}:
Correctness is decided by the ability of tolerating the intra-user biometric variation to generate a deterministic key.
Thus, we analyze the variation tolerability of IECO with respect to its parameters, then, the correctness could be inferred. 

Let $\omega$ and $\omega'$ be the strings extracted from the same user.
Let $\zeta$ be the probability of a bit $i$ to be error (\textit{i.e.,} $\mathsf{P}_{(\omega'_i\neq\omega_i)}=\zeta$).
For simplicity, we assume the errors can independently occur in any position with the same probability.

From the IECO's construction, the key $\kappa$ is successfully reproduced if $m'=m$ (assuming that the error of DL is negligible, \textit{i.e.,} $2^{-\gamma}\approx 0$).
This can be satisfied if $\mathsf{d}_H(c',c)\leq t$. 
Thus, we can firstly see that, the variation tolerability is decided by the adopted ECC.
On the other hand, $c_i$ and $c'_i$ can be different in two cases: 
\begin{enumerate}
	\item[\textit{(i)}]
	\textit{$c_i=1, c'_i=0$}: 
	This occurs when $s_i\neq s'_i$ and $c_i=1$ (\textit{i.e.,} the position $i$ that $s_i$ is used to create $p_i$ instead of using a random $r_i$). 
	As each symbol $s_i$ is formed from $\phi$ bits of $\omega$, the probability of this case is
	\begin{equation}\label{eq_prob_1}
	\mathsf{P}_{(c'_i\neq c_i|c_i=1)}=\mathsf{P}_{(s'_i\neq s_i|c_i=1)}=\frac{1}{2}\phi\zeta,
	\end{equation}
	(assuming that the numbers of bit 1s and 0s in $c$ are equal). 

	\item[\textit{(ii)}]
	\textit{$c_i=0, c'_i=1$}: 
	For the position $i$ having $c_i=0$, an incorrect $c'_i$ is returned if $s'_i=r_i$, where $r_i$ is the random symbol generated in the key generation phase. 
	The probability for this case is 
	\begin{equation}\label{eq_prob_2}
	\mathsf{P}_{(c'_i \neq c_i|c_i=0)} = \mathsf{P}_{(s'_i = r_i|c_i=0)}=\frac{1}{2}\frac{\phi\zeta}{(2^\phi-1)}.
	\end{equation}
\end{enumerate} 
From \eqref{eq_prob_1} and \eqref{eq_prob_2}, the probability of $s_i$ to be different from $s'_i$ is 
\begin{equation}\label{eq_intra_err_with_phi}
\mathsf{P}_{(c'_i\neq c_i)}=\frac{1}{2}(\phi\zeta+\frac{\phi\zeta}{(2^\phi-1)})
		=\frac{1}{2}\zeta(\phi+\frac{\phi}{2^\phi-1}).
\end{equation}
We could see that, with $\zeta$ as a constant, then $\mathsf{P}_{(c'_i\neq c_i)}$ is represented as an increasing function with respect to $\phi\in \mathbb{N}^*$. 
So, under the same variation condition of biometric data (specified by $\zeta$), increasing $\phi$ will increase the probability of error between $c$ and $c'$, thus, reduce the model's correctness.

\textbf{Secureness}: 
We measure the impacts of IECO's parameters on secureness.
Let $\omega^u$ and $\omega^v$ be the strings extracted from users $u$ and $v$, and $\kappa$ be a key generated from $\omega^u$.
We analyze the probability of successfully reproducing $\kappa$ using $\omega^v$, with respect to the IECO's parameters.
We assume that $\mathsf{P}_{(\omega^u_{i}=\omega^v_{i})}=\eta$, $\forall i=1..\phi n$, where $\omega^u_{i}$ and $\omega^v_{i}$ are the bits $i$ of $\omega^u$ and $\omega^v$, respectively.

We could see that, user $v$ can successfully obtain $\kappa$ if $\mathsf{d}_{H}(c^u, c^v)\leq t$, where $c^u$ is the codeword generated in the enrollment phase of user $u$, and $c^v$ is the codeword reproduced by $\omega^v$.
Thus, the secureness is firstly decided by the error correcting capability $t$ of the ECC, in which, increasing $t$ will reduce the secureness.
Next, we consider the impact of $\phi$. There are two cases for $c^u_{i}=c^v_{i}$:
\begin{enumerate}
	\item[\textit{(i)}]
	$c^u_{i}=1,c^v_{i}=1$: This happens when $s^v_{i}=s^u_{i}$ and $c^u_{i}=1$. 
	As each symbol is formed from $\phi$ bits of $\omega$,
	the probability of this case is
	\begin{equation}
	\mathsf{P}_{(c^v_{i}=c^u_{i}|c^u_{i}=1)}=\mathsf{P}_{(s^u_{i}=s^v_{i}|c^u_{i}=1)}=\frac{1}{2}\eta^\phi.
	\end{equation} 
	\item[\textit{(ii)}] 
	$c^u_{i}=0,c^v_{i}=0$: This happens when $c^v_{i}=0$ and $s^v_{i}=r_i$, where $r_i$ is the random symbol generated in the key generation phase.
	The probability of this case is
	\begin{equation}
	\begin{split}
		\mathsf{P}_{(c^v_{i}=c^u_{i}|c^u_{i}=0)}=\mathsf{P}_{(s^v_{i}\neq r_i|c^u_{i}=0)}
		=\frac{1}{2}(\eta^\phi+(1-\eta^\phi)\frac{2^\phi-2}{2^\phi-1}).
	\end{split}
	\end{equation}
\end{enumerate}

In summary, the probability for $c^u_{i}$ to be equal to $c^v_{i}$ is 
\begin{equation}
\mathsf{P}_{(c^v_{i}=c^u_{i})}= \eta^\phi + \frac{1}{2}(1-\eta^\phi)\frac{2^\phi-2}{2^\phi-1}.
\end{equation}
We can see that $\mathsf{lim}_{\phi \to \infty}\mathsf{P}_{(c^v_{i}=c^u_{i})}=0.5$.
That means, when increasing the symbol size $\phi$, the probability of a bit $c^v_{i}$ to be equal to $c^u_{i}$ will approach 0.5, thus, it is more difficult a user to reproduce the key generated by another user. 

In conclusion, the model's correctness and secureness are controlled by:
\begin{itemize}
	\item
	\textit{Error correcting capability $t$:} Increasing $t$ will increase the correctness, however, decrease the model's secureness. 
	\item 
	\textit{Symbol size $\phi$:} Increasing $\phi$ will lead to a more secure model, however, decrease the model's correctness. 
\end{itemize}

\subsubsection{Security and Privacy Improvement}\label{sssec_security_improve}
Here, we describe the security improvement over the original scheme.

\textbf{Weakness of ECO}: The ECO scheme could not ensure irreversibility, thus, it is a non-reusable scheme.
Specifically, with the construction described in \ref{ssec_eco_scheme}, we assume that the key $m$ is compromised.
Then, the attacker could determine the used codeword $c$ as $c=\mathsf{encode}_\mathcal{C}(m)$.
By observing $c$, he can identify which locked points $p_i$ are derived from the biometric symbols. 
Subsequently, given a locked point $p_i$, the sealed symbol $s_i$ could be identified with the computational cost as $2^\phi$. 
As $\phi$ is usually small, $s_i$ could be effectively obtained by exhaustively searching all possible values in $\mathcal{Z}$.
When sufficient components of $\mathbf{s}$ are known, the attacker can compromise all other keys generated from the same user.

\textbf{Our improvement}: In IECO, instead of using $m$ as the generated key, we use $m$ to lock $\kappa$ with a DL. 
$\kappa$ is output as the generated key while $m$ is discarded.
In case of $\kappa$ is compromised, $L_\kappa$ and $\kappa$ provide no information to compute $m$ \cite{canetti2010symmetric}.
Guessing $m$ from $\kappa$ and $L_\kappa$ is equivalent to brute force all possible values of $m$.
Thus, when $\kappa$ is compromised, the adversary gains no advantage in reconstructing the biometric data.

\section{Experimental Evaluation}\label{sec_expe}

In this section, we present the experimental evaluations for the proposed model.
First, we explain the datasets and experiment procedure, then, report the performance under optimal parameters.
In addition, we analyze the adopted techniques in details, and present the evaluations on security and user privacy. 
Finally, a relative comparison to existing researches is provided.

\subsection{Experiment Procedure}\label{ssec_dataset}
Our model was evaluated with OU-ISIR \cite{ngo2014largest} and whuGAIT datasets \cite{zou2020deep}. 

\subsubsection{OU-ISIR Dataset}

OU-ISIR is known as the largest population public gait dataset, formed from the IMU-based gait data of 744 users. 
We divided this dataset into two non-overlapping sets for training and testing.
The training set consisted of $520$ users, and the testing set contained data of other $224$ users.

Data sequences in the training set were split into segments of $w=100$ signals.
Two segments of each user formed a validating set $\mathcal{V}$, and all remaining segments formed a learning set $\mathcal{L}$.
Note that, two consecutive segments in $\mathcal{L}$ overlapped each other $97\%$.
$\mathcal{L}$ is used to train the network from scratch, with the mini-batch size of $128$. 
The CNN branch was updated using Adam algorithm with the learning rate as $15\times 10^{-6}$.
For the LSTM branch, Stochastic Gradient Descent algorithm with the learning rate of $0.15$ and momentum of $0.9$ was used.
When completing an epoch, the validation loss over the set $\mathcal{V}$ was computed. 
If this loss did not decrease during $15$ epochs, we terminated the training process.
Then, the network was used as a feature extraction tool to extract representation template from the gait segment.

Data of 224 users in the testing set were used to build and evaluate 224 key generation models (one model for each user).
Specifically, for each user $u$, his/her data were divided into two non-overlapping and equal parts.
The gait segments from the first part were used as the enrolled data to generate a key $\kappa_u$.
Then, the remaining data of $u$, along with data of $223$ other users, were used to reproduce $\kappa_u$.
To increase the stability of the reproduced key, we used $5$ segments for each attempt. 
Specifically, given $5$ gait segments, $5$ representation templates $\mathbf{f}^i$ were extracted.
Then, their mean template was determined and used to reproduce the key.
The performance was evaluated with FRR and FAR.
Let $N_u$ be the number of times a user tried to reproduce his generated key, and $F_u$ be the failed times. 
The FRR was determined as 
$\text{FRR}=\frac{F_u\times 100}{N_u}$.
Let $N_o$ be the number of times that the data of other users were used to reproduce the key, and $F_o$ be the success times.
The FAR was estimated as $
\text{FAR} = \frac{F_o \times 100}{N_o}$.

\subsubsection{WhuGAIT Dataset}
WhuGAIT dataset consists of IMUs-based gait data of $118$ users, acquired in realistic conditions.
We also divided this dataset into two non-overlapping sets for training and testing.
The training set consisted of $98$ users, and the testing set contained data of the remaining $20$ users.
Then, the experiment procedure with this dataset was conducted similarly to the OU-ISIR dataset.

\subsubsection{Parameter Fine-tuning}
In \eqref{eq_final_loss}, $\alpha$ and $\beta$ need to be fine-tuned for optimal performance. 
Exhaustively trying all of their combinations is a computationally expensive task.
Thus, we used an adaptive search that manually examined the output of a combination and adjusted the parameters accordingly.
We started by training the model with $\alpha=\beta=1$. %
The reliable strings $\omega$ were extracted and the histogram of normalized Hamming distance was estimated (as in Figure \ref{fig_hamd_reliable_str}).
If the Hamming distance of intra-class was high, we increased the value of $\alpha$ by $0.1$, while keeping $\beta$ unchanged.
By changing $\alpha$, we could find a good trade-off between intra-class stability and inter-class separability.
Then, we adjusted $\beta$ ($0.1$ in each iteration) to balance the randomness and intra-class stability.
By increasing $\beta$, the histogram of inter-class normalized Hamming distance would centralize to $0.5$ harder, however, the intra-class Hamming distance would increase. 
 
By the above process, we found that, the model achieved the best balance with $\alpha=1.2$, $\beta=0.9$ for OU-ISIR, and $\alpha=1.1$, $\beta=1.8$ for whuGAIT.
%

\subsection{Overall Performance}\label{ssec_ov_per}

Our scheme achieved the best performance when using IECO with BCH code of length $255$ and symbol size $\phi=2$.
The detailed performances at different key sizes are summarized in Table \ref{tab_overall_result} (on the assumption of ideal DL and OPF, \textit{i.e.,} $2^{-\gamma}\approx 0$). 
We could see that, under the same settings of codeword length and symbol size, the key size $k$ controls the trade-off between the FAR (\textit{i.e.,} secureness) and FRR (\textit{i.e.,} correctness).
Increasing $k$ will decrease the error correcting capability $t$ of the code $\mathcal{C}$, thus 
the FAR is reduced and FRR is increased.
This confirms the analysis in \ref{sssec_param_analy}.

\begin{figure*}[t]
	\centering
	\begin{subfigure}{.5\textwidth}
		\centering
		\includegraphics[scale=0.32]{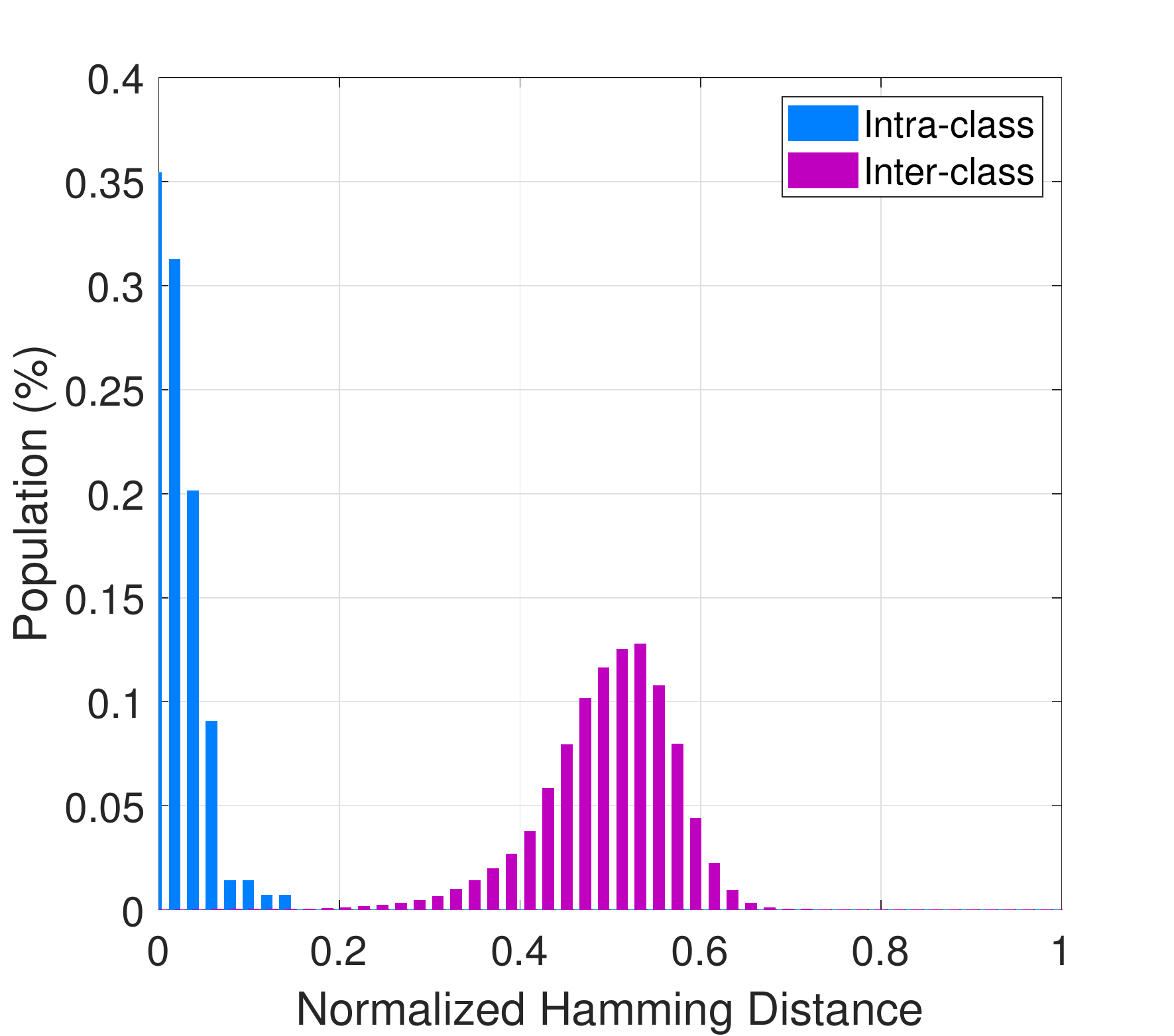}
		\caption{OU-ISIR dataset}
		\label{fig_hamd_eico_ou}
	\end{subfigure}%
	\begin{subfigure}{.5\textwidth}
		\centering
		\includegraphics[scale=0.32]{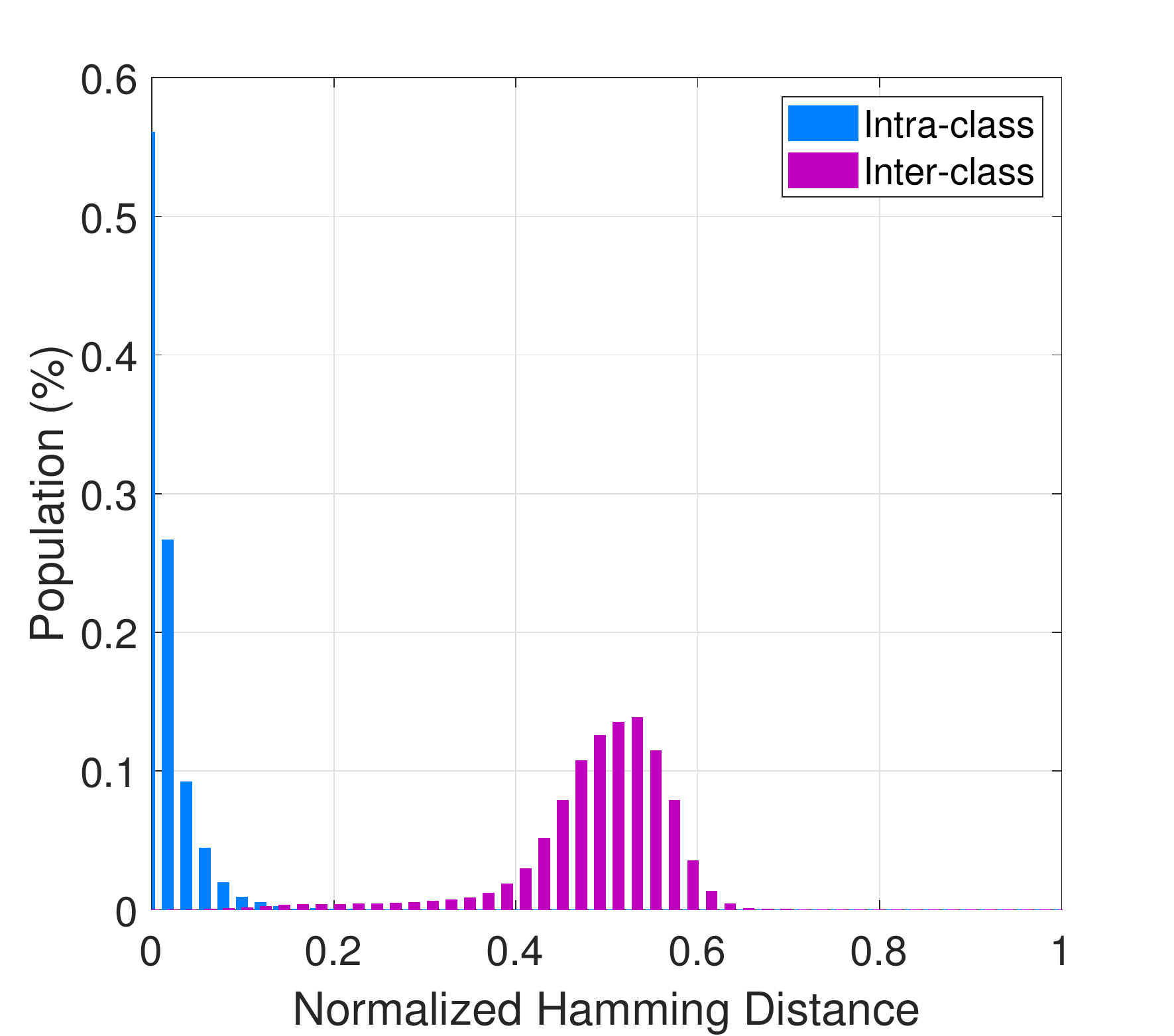}
		\caption{WhuGAIT dataset}
		\label{fig_hamd_eico_whu}
	\end{subfigure}%
	\caption{The normalized Hamming distance between the enrolled codeword $c$ and the reproduced codewords $c'$.} 
	\label{fig_hamd_eico}
\end{figure*}

Figures \ref{fig_hamd_eico} show the normalized Hamming distance between the enrolled codeword $c$ and the reproduced codewords $c'$. 
We could see that, when $c'$ is reproduced by the enrolled user (intra-class), it showed low variation from $c$. 
On the other hand, when using the other users' data (inter-class), the distance between $c'$ and $c$ follows a Gaussian of mean approximated to $0.5$.

\begin{table}[t]
	\centering
	\caption{Performances at different key sizes when using BCH code of length $255$ and symbol size $\phi=2$.}
	\label{tab_overall_result}
	\def\arraystretch{1.1}
	
	\begin{tabular}{cccccc} \hlineB{3}
		\textbf{\textit{Key size}}  & \textbf{\textit{Error}}& \multicolumn{2}{c}{\textbf{\textit{OU-ISIR}}} & \multicolumn{2}{c}{\textbf{\textit{whuGAIT}}} \\
		(bits)&\textit{\textbf{Tolerability}}& $FAR (\%)$ & $FRR (\%)$& $FAR (\%)$ & $FRR (\%)$\\
		\hlineB{2}
		$115$ & 	$21$		& $0.001$ 				& $2.083$ 		&$0.022$ 	& $2.177$ \\ 
		$123$ & 	$19$		& $0$ 					& $4.167$  		&$0.005$	& $2.66$ \\ 
		$131$ & 	$18$		& $0$ 					& $4.167$  		&$0$		& $3.265$\\ 
		$139$ & 	$15$		& $0$ 					& $4.167$  		&$0$		& $5.441$\\ 
		$147$ & 	$14$		& $0$ 					& $6.25$   		&$0$		& $6.771$\\ \hlineB{3}
	\end{tabular}
\end{table}

\subsection{Discussion}\label{ssec_discussion}

In this section, we provide detailed analyses for the adopted methods, to confirm their impacts on the overall performance and security.

\subsubsection{Optimization Function}\label{sssec_loss_function_analyze}
\begin{figure*}[t]
	\centering
	\begin{subfigure}{.34\textwidth}
		\centering
		\includegraphics[scale=0.22]{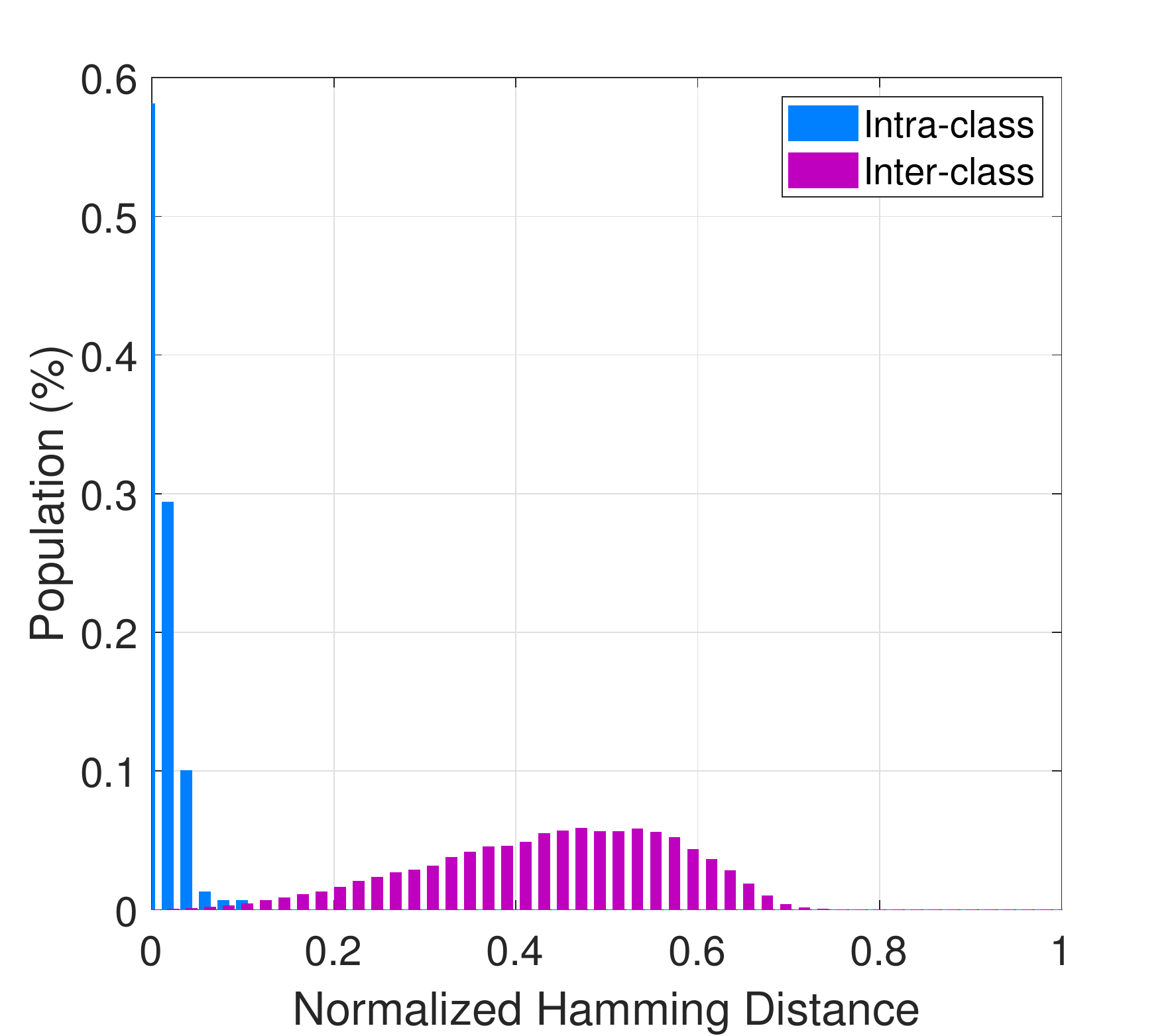}
		\caption{The original triplet loss \cite{schroff2015facenet}.}
		\label{fig_hamd_reliable_str_tlloss}
	\end{subfigure}%
	\begin{subfigure}{.34\textwidth}
		\centering
		\includegraphics[scale=0.22]{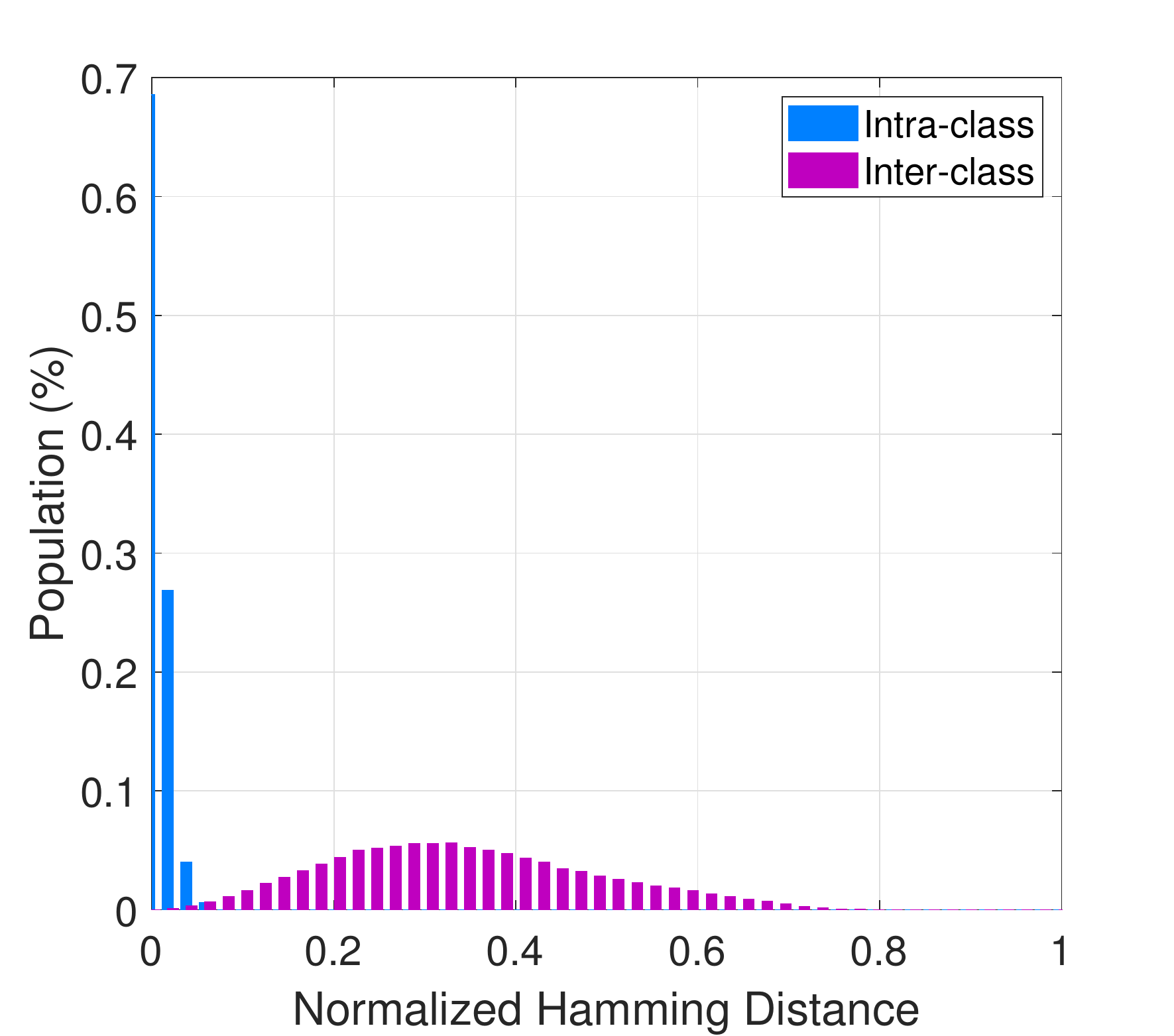}
		\caption{The loss proposed in \cite{talreja2020deep}}
		\label{fig_hamd_reliable_str_trpl_sta}
	\end{subfigure}%
	\begin{subfigure}{.34\textwidth}
		\centering
		\includegraphics[scale=0.22]{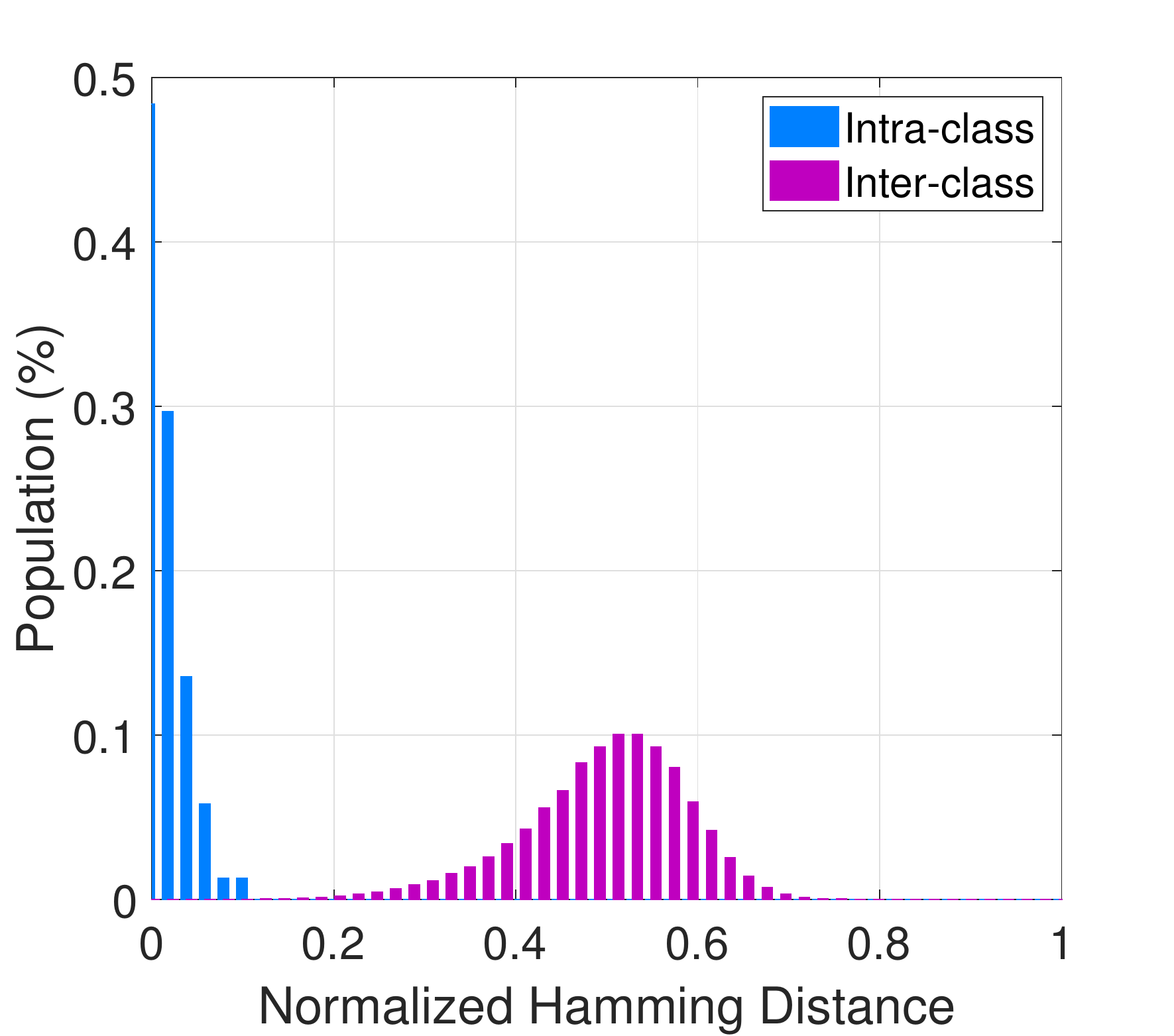}
		\caption{Our proposed loss function.}
		\label{fig_hamd_reliable_str}
	\end{subfigure}%
	\caption{The normalized Hamming distance of the reliable binary strings $\omega$ when using our loss function in comparing to existing solutions.}
	\label{fig_hamming_comparing}
\end{figure*}

First, we compare our loss function with existing solutions (\textit{i.e.,} \cite{schroff2015facenet,talreja2020deep}). 
Figures \ref{fig_hamming_comparing} show the normalized Hamming distance of the reliable string $\omega$ extracted with different loss functions, measured on OU-ISIR dataset.
With triplet loss \cite{schroff2015facenet}, the strings extracted from a same user are highly stable, \textit{i.e.,} the normalized intra-class Hamming distance distribution approximately follows a half-Gaussian of mean $\mu=0$ and standard deviation $\sigma=0.0156$ (Figure \ref{fig_hamd_reliable_str_tlloss}). 
However, this loss function does not address the inter-class randomness (\textit{i.e.,} inter-class Hamming distance follows a Gaussian of $\mu=0.4365$ and $\sigma=0.1703$).
Poor inter-class randomness would cause the model to be vulnerable from several attacks (\textit{e.g.,} statistical attack, nearest impostor attack) \cite{stoianov2009security}.

The study \cite{talreja2020deep} proposed two constraints to improve the randomness and stability of the extracted binary string.
The first one directs each feature to get closer to $1$ or $-1$, thus, increases its stability, 
$L_2=\frac{1}{N}\sum_{i=1}^{|\mathcal{B}|}\|\mathbf{f}^i\|$
where $|\mathcal{B}|$ is the mini-batch size, and $N$ is the length of the feature template.
The second constraint aims to balance the numbers of bit $1$ and $0$ in the extracted string, 
$L_3=\frac{1}{N}\sum_{i=1}^{|\mathcal{B}|} \sum_{j=1}^N f_j^i,$ 
where $f_j^i$ is the feature $j$ of template $i$.
Figure \ref{fig_hamd_reliable_str_trpl_sta} plots the normalized Hamming distance of $\omega$ when combining these constraints with triplet loss.
Comparing to using only triplet loss, incorporating these constraints improves the intra-class stability (\textit{i.e.,} $\sigma$ is reduced to $0.0108$). 
However, the inter-class separability and randomness are not improved. 

With our loss function, although the stability of intra-class is minorly reduced (\textit{i.e.,} the half-Gaussian for intra-class has $\mu\approx0$ and $\sigma=0.0251$), the inter-class randomness is remarkably increased (\textit{i.e.,} $\mu=0.496$ and $\sigma=0.0871$).

\begin{figure*}[t]
	\centering
	\begin{subfigure}{.33\textwidth}
		\centering
		\includegraphics[scale=0.17]{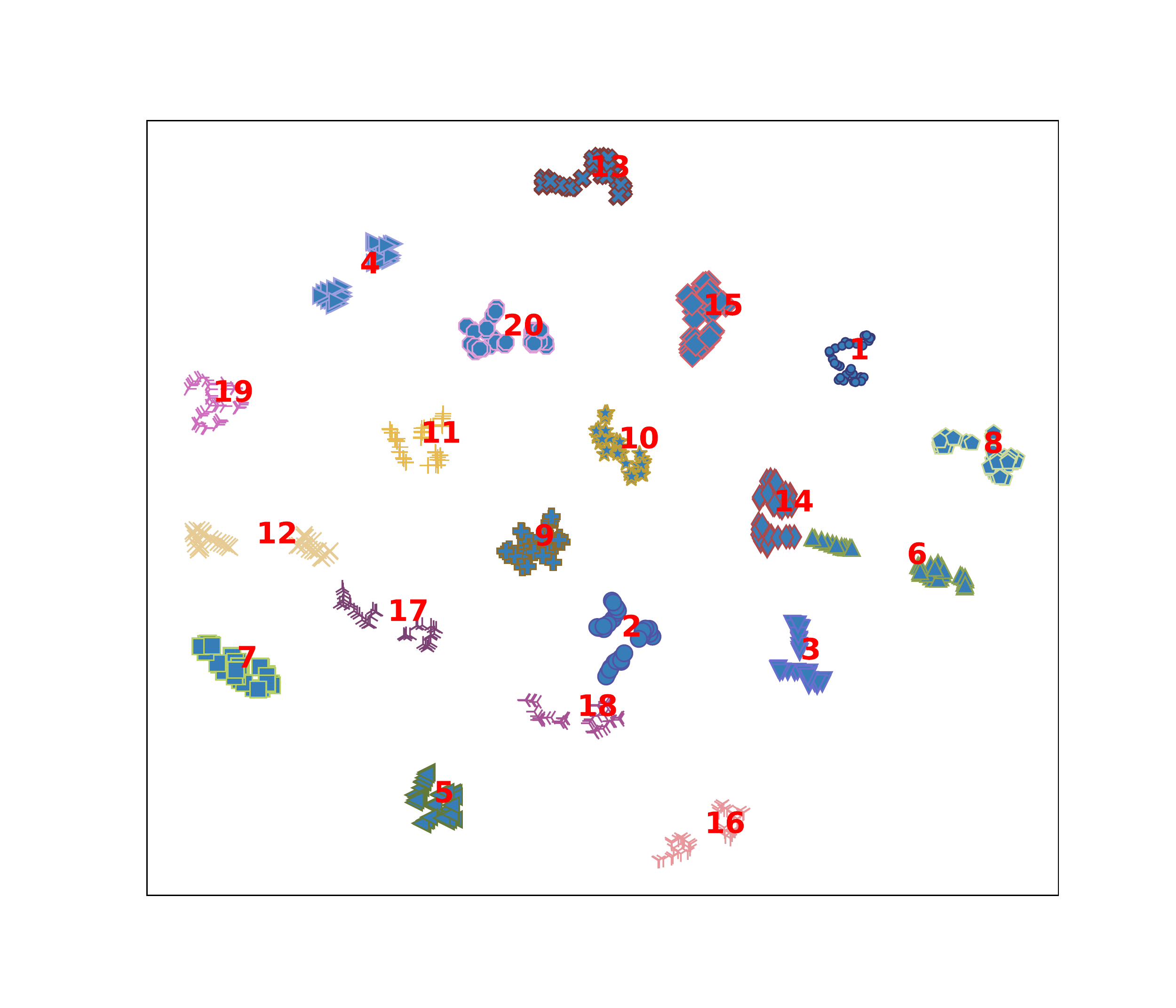}
		\caption{The original triplet loss \cite{schroff2015facenet}.}
		\label{fig_tsne_triplet}
	\end{subfigure}%
	\begin{subfigure}{.33\textwidth}
		\centering
		\includegraphics[scale=0.17]{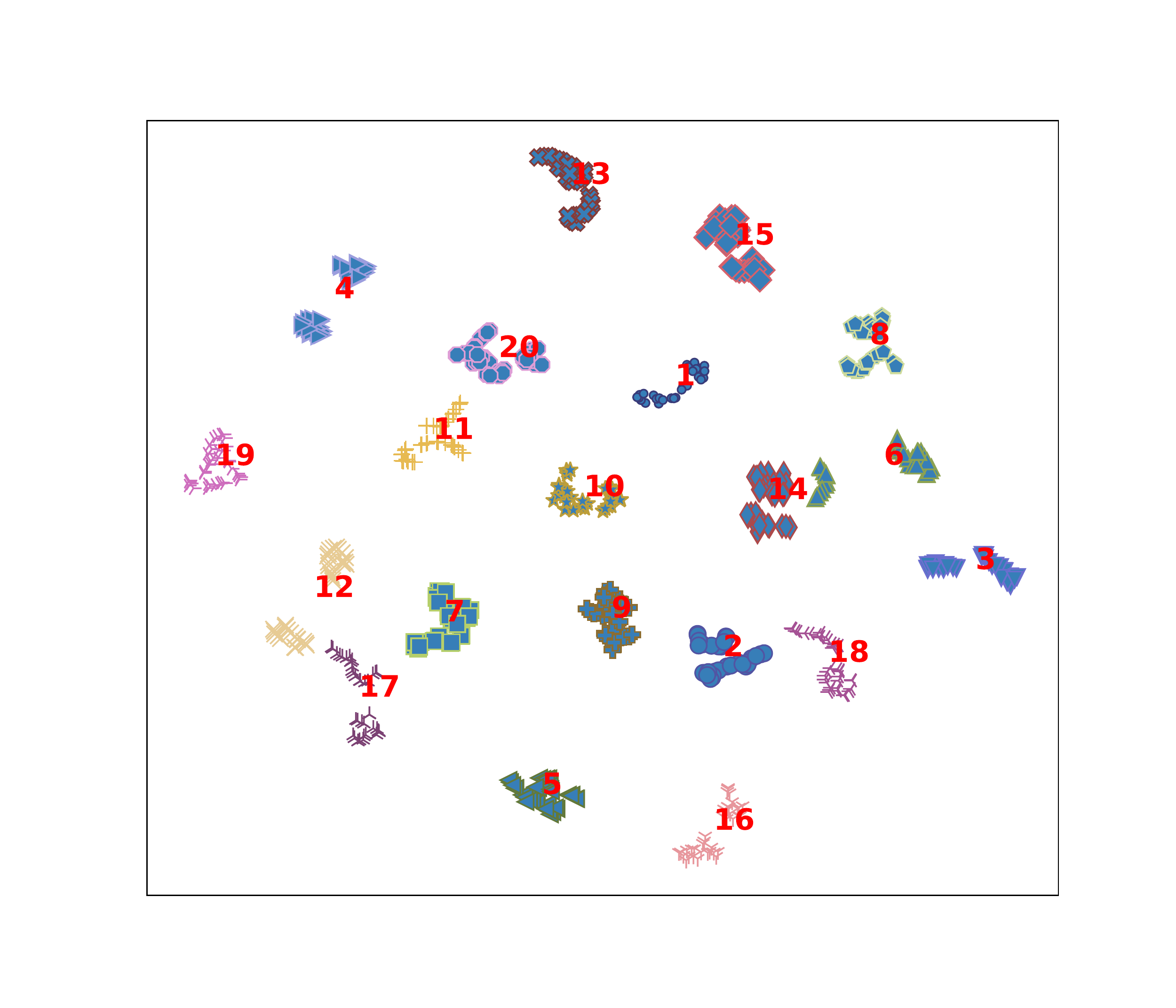}
		\caption{The loss proposed in \cite{talreja2020deep}}
		\label{fig_tsne_triplet_tifs}
	\end{subfigure}%
	\begin{subfigure}{.33\textwidth}
		\centering
		\includegraphics[scale=0.17]{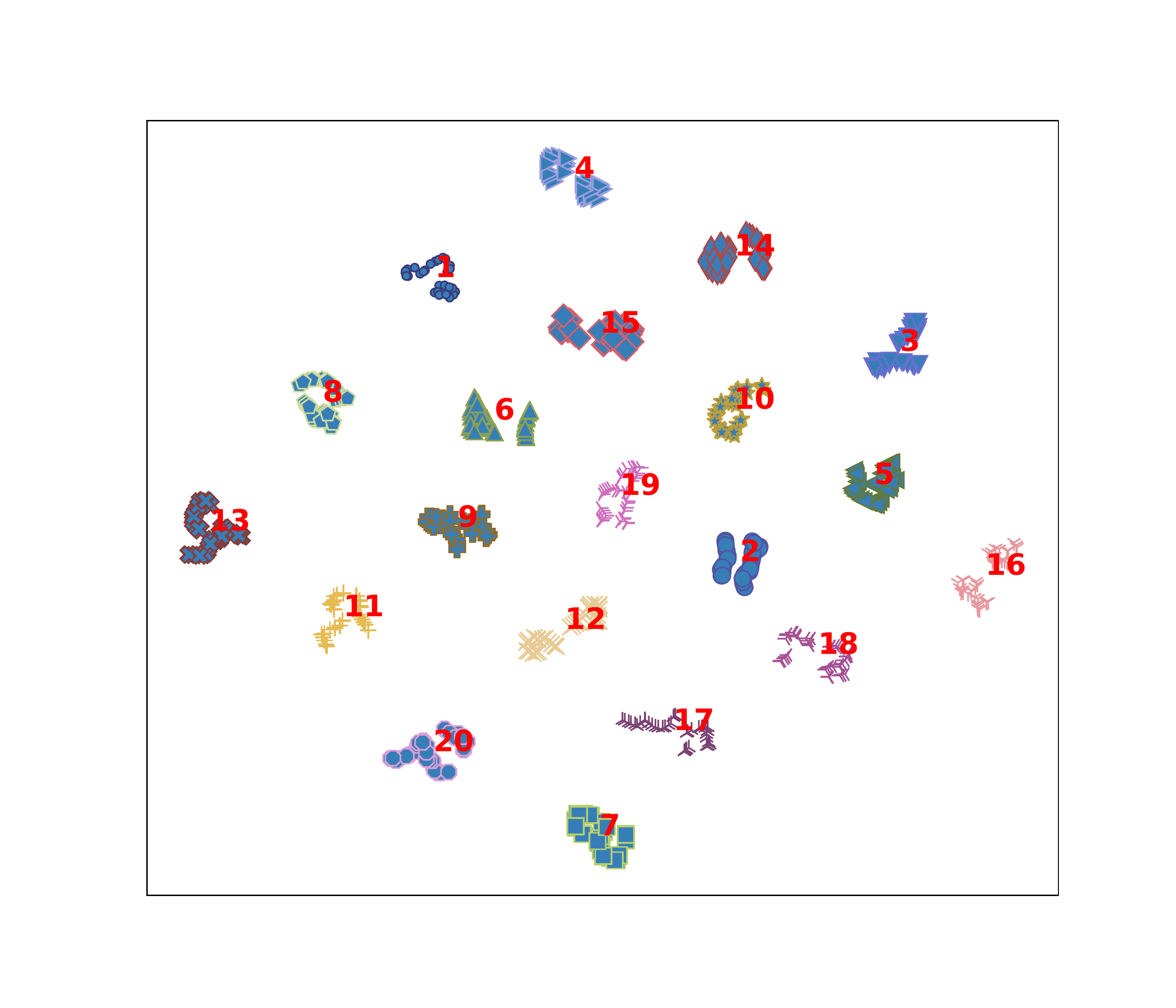}
		\caption{Our proposed loss function.}
		\label{fig_tsne}
	\end{subfigure}%
	\caption{T-SNE visualization of the extracted feature $\mathbf{f}$ of $20$ testing users in OU-ISIR dataset.}
	\label{fig_tsne_comparing}
\end{figure*}

Figures \ref{fig_tsne_comparing} plot the visualization of the representation vectors $\mathbf{f}$ using t-Distributed Stochastic Neighbor Embedding (t-SNE) \cite{van2008visualizing}.
For the loss functions \cite{schroff2015facenet, talreja2020deep}, although the representation vectors are grouped into separated clusters, some clusters still close to each other (\textit{e.g.}, $2$ and $18$; $14$ and $6$; $12$ and $17$).
With our loss function, the clusters are clearly separated from each other by a certain margin, and no cluster is too close to another.

\subsubsection{Symbol size}\label{sssec_symbol_size}
\begin{figure*}[t]
	\centering
	\begin{subfigure}{.5\textwidth}
		\centering
		\includegraphics[scale=0.32]{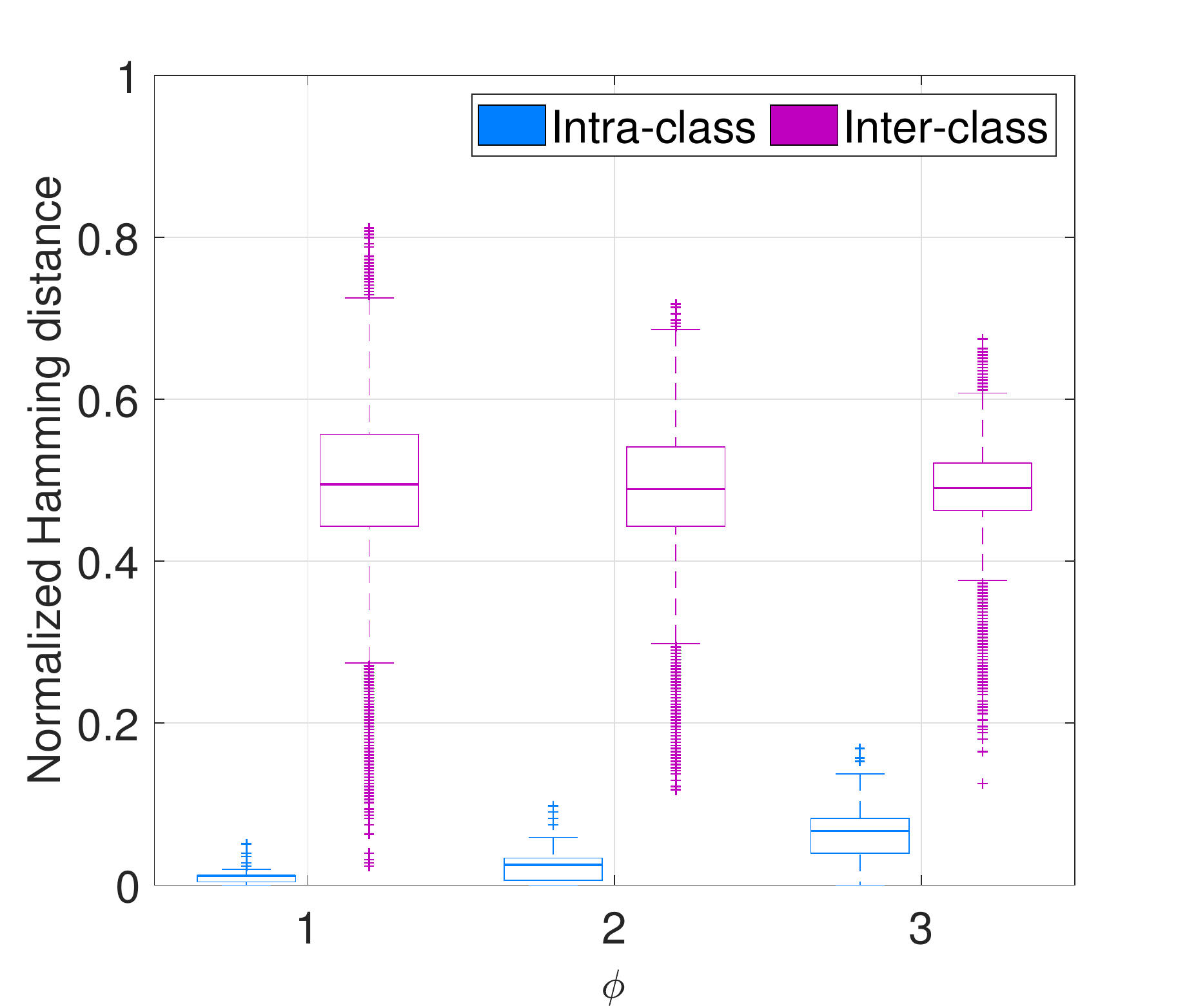}
		\caption{OU-ISIR dataset}
		\label{fig_ham_dif_symbol_size_ou}
	\end{subfigure}%
	\begin{subfigure}{.5\textwidth}
		\centering
		\includegraphics[scale=0.32]{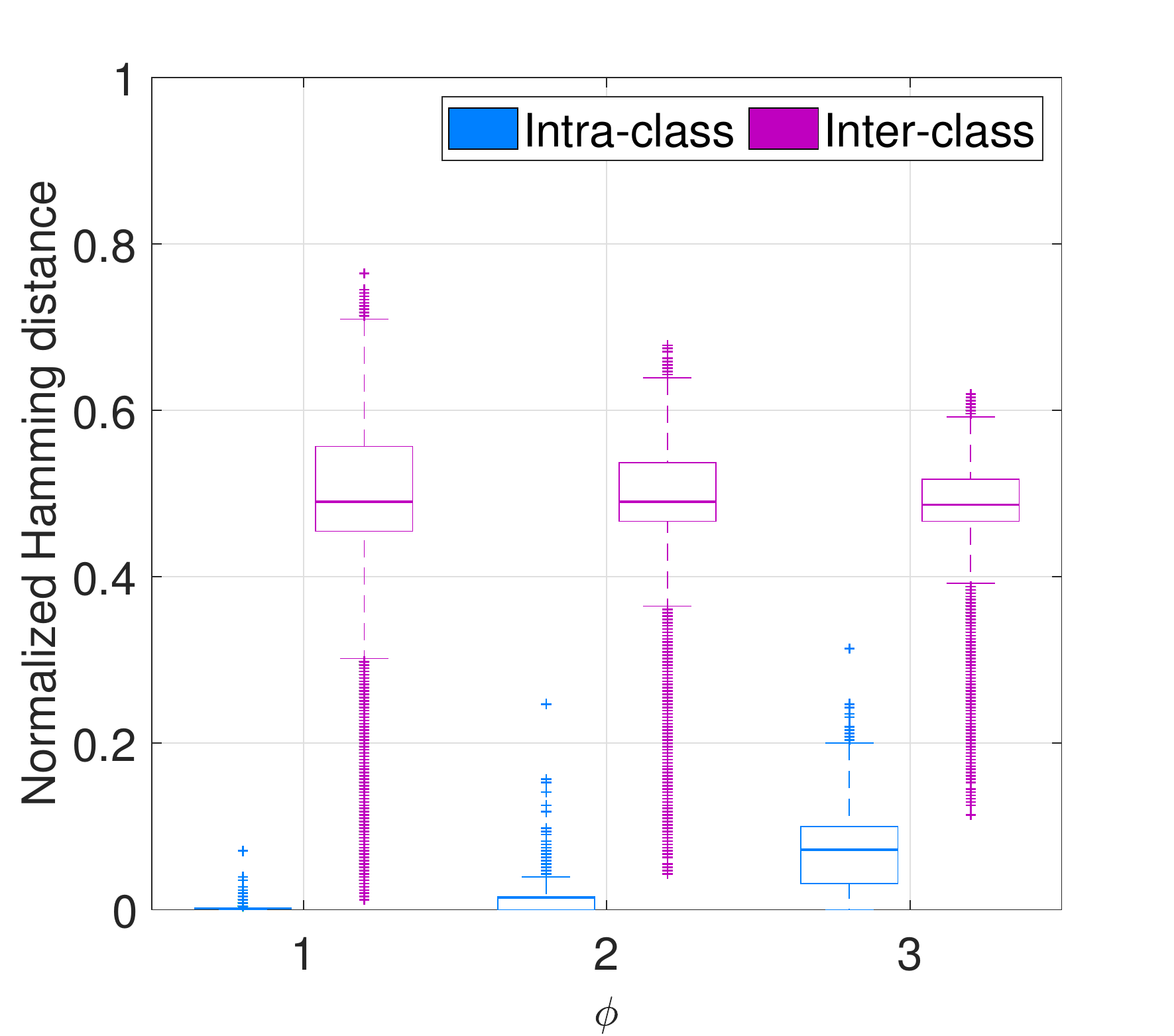}
		\caption{WhuGAIT dataset}
		\label{fig_ham_dif_symbol_size_whu}
	\end{subfigure}%
	\caption{The normalized Hamming distance distribution measured on $c$ under different settings of symbol size $\phi$.}
	\label{fig_ham_dif_symbol_size}
\end{figure*}

We experimentally analyzed the impact of symbol size $\phi$ on the overall performance and security. 
Figures \ref{fig_ham_dif_symbol_size} display the Hamming distance between the enrolled codeword $c$ and the reproduced codeword $c'$ when using different symbol sizes.
We observed that, when $\phi$ is increased, the intra-class Hamming distance is also increased, which means the error toleration capability is reduced.
Meanwhile, increasing $\phi$ makes the inter-class Hamming distance to be harder centralized around the mean $0.5$, thus, increases the security.
These observations confirm the theoretical analysis of $\phi$ in \ref{sssec_param_analy}.

\subsection{Privacy and Security Analyses}\label{ssec_secu_pri_ana}

As specified in ISO/IEC standard \cite{information2011iso}, irreversibility and unlinkability are mandatory for a biometric template protection solution. 
In this section, the evaluations on irreversibility and unlinkability are provided. %
In addition, we analyze the model security against practical attacks on BCS solutions.
\subsubsection{Irreversibility}
Irreversibility refers to the computational hardness of inferring biometric data from the extracted helper data or/and the generated key.

First, we consider a scenario in which the extracted key $\kappa$ has not been compromised, and the attacker tries to infer the biometric features from the extracted helper data.
All the helper data include the deep network's weights and biases, the RP matrix $\mathbf{R}$, the reliable feature index, and the locked points $\{\mathcal{P}, L_\kappa\}$. 
The network is not trained by the enrolled user's data, and the representative template is discarded after enrollment.
Thus, it is impractical to infer the enrolled user's data from the network \cite{osadchy2018all}.
The RP matrix is randomly generated, and is independent from the gait data (see \ref{ssec_revoca_bin_block}), thus, it provides no information to infer the gait data.
Although the reliable feature index shows the reliability order of extracted features, it is useless for guessing the actual value of gait features. 
From each locked point $p_i \in \mathcal{P}$, the attacker can infer the inside plaintext with the computational complexity as $2^\phi$. 
As $\phi$ is a small number (\textit{i.e.,} $\phi=2$ in our model), the value contained inside $p_i$ can be obtained easily.
However, without $c_i$ (which is unknown from the attacker), the attacker can not determine whether $p_i$ is the locked point of a biometric symbol $s_i$ or a random string $r_i$. 
There are $n=255$ locked points, where $n$ is the codeword length.
Thus, the complexity of guessing the biometric template given all locked points is $2^{255}$.
$L_\kappa$ is the locked point of $\kappa$ using the key as $m$, (\textit{i.e.,} $L_\kappa\leftarrow \mathsf{dLock}(m, \kappa)$).
Given only $L_\kappa$, it is computationally impractical to obtain $m$ \cite{canetti2010symmetric}.

Next, we consider a scenario in which the key $\kappa$ has been compromised (\textit{e.g.,} user may mistakenly submit $\kappa$ to a fake application).
Given $\kappa$ and $L_\kappa$, the only way for the attacker to obtain $m$ is guessing all possible values of $m$ \cite{canetti2010symmetric}. 
The complexity of this task is $2^k=2^{139}$, where $k$ is the message length of the adopted BCH code, which is also the model's security level.
Thus, in our model, even the generated key $\kappa$ has been compromised, the attacker still can not gain any additional advantage on reconstructing the original gait template.

\subsubsection{Unlinkability}
We adopted the framework proposed in \cite{gomez2017general} to evaluate the unlinkability.
%
In this method, the \textit{mated} and \textit{non-mated} scores will be evaluated, to measure the similarity of the extracted strings when changing only the parameters (\textit{i.e.}, RP matrix), and when changing both the parameters and the biometric source.
Then, two metrics are computed to evaluate the unlinkability. 
The first one is \textit{local score-wise}, $\text{D}_{\leftrightarrow}{(s)}\in[0,1]$, which reflects the likelihood ratio of mated and non-mated score distribution in a specific score. 
In a specific score,  $\text{D}_{\leftrightarrow}{(s)}=1 $ means the system is fully linkable, while $\text{D}_{\leftrightarrow}{(s)}=0$ implies the fully unlinkability. 
The second one is \textit{global measure}, $\text{D}_{\leftrightarrow}^{sys} \in [0, 1]$, gives an overall quantification of the unlinkability. 
For a fully unlinkable model, 
$\text{D}_{\leftrightarrow}^{sys}$ should be close to zero.

\begin{figure*}[t]
	\centering
	\begin{subfigure}{.5\textwidth}
		\centering
		\includegraphics[scale=0.30]{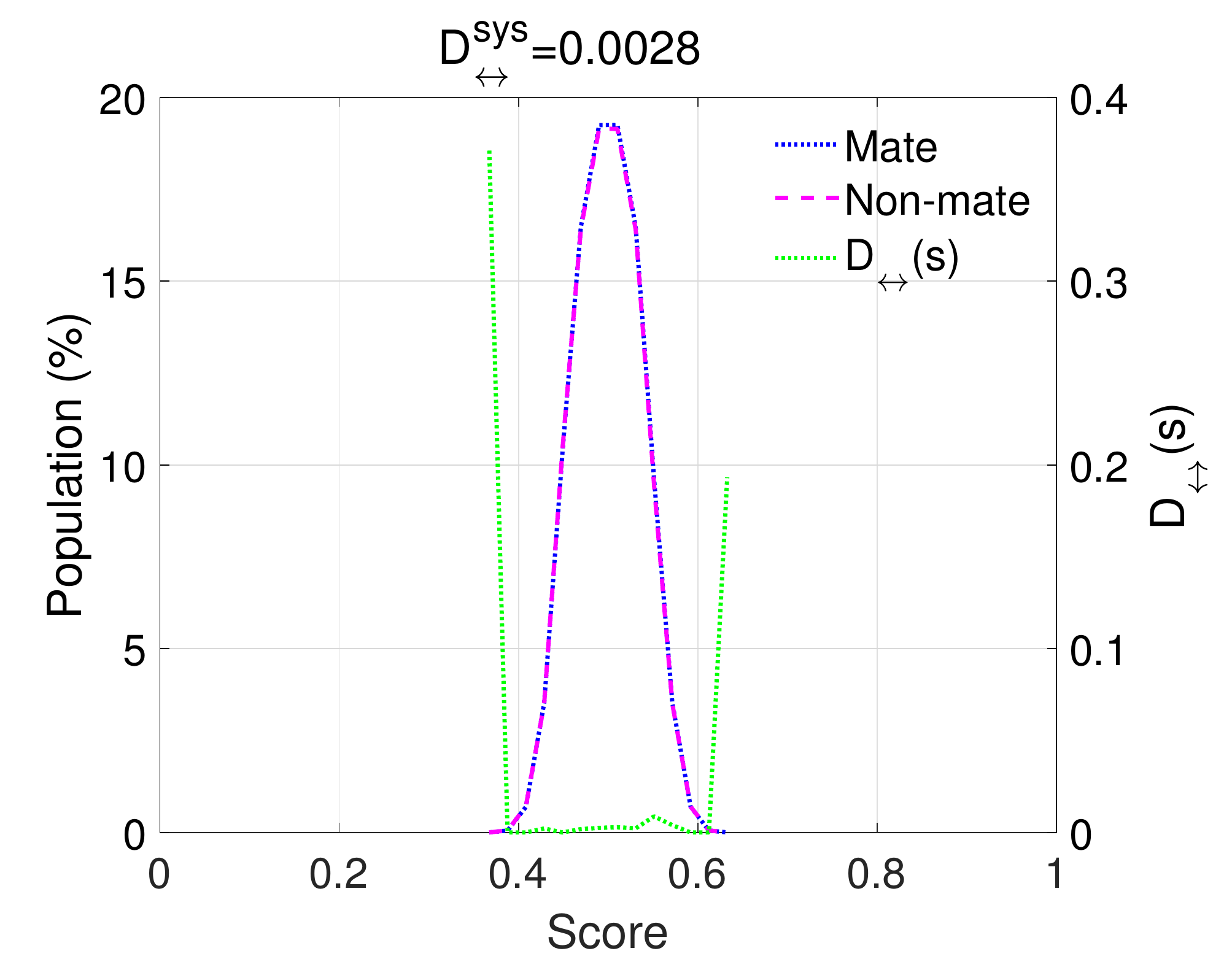}
		\caption{OU-ISIR dataset}
		\label{fig_unlink_ana}
	\end{subfigure}%
	\begin{subfigure}{.5\textwidth}
		\centering
		\includegraphics[scale=0.30]{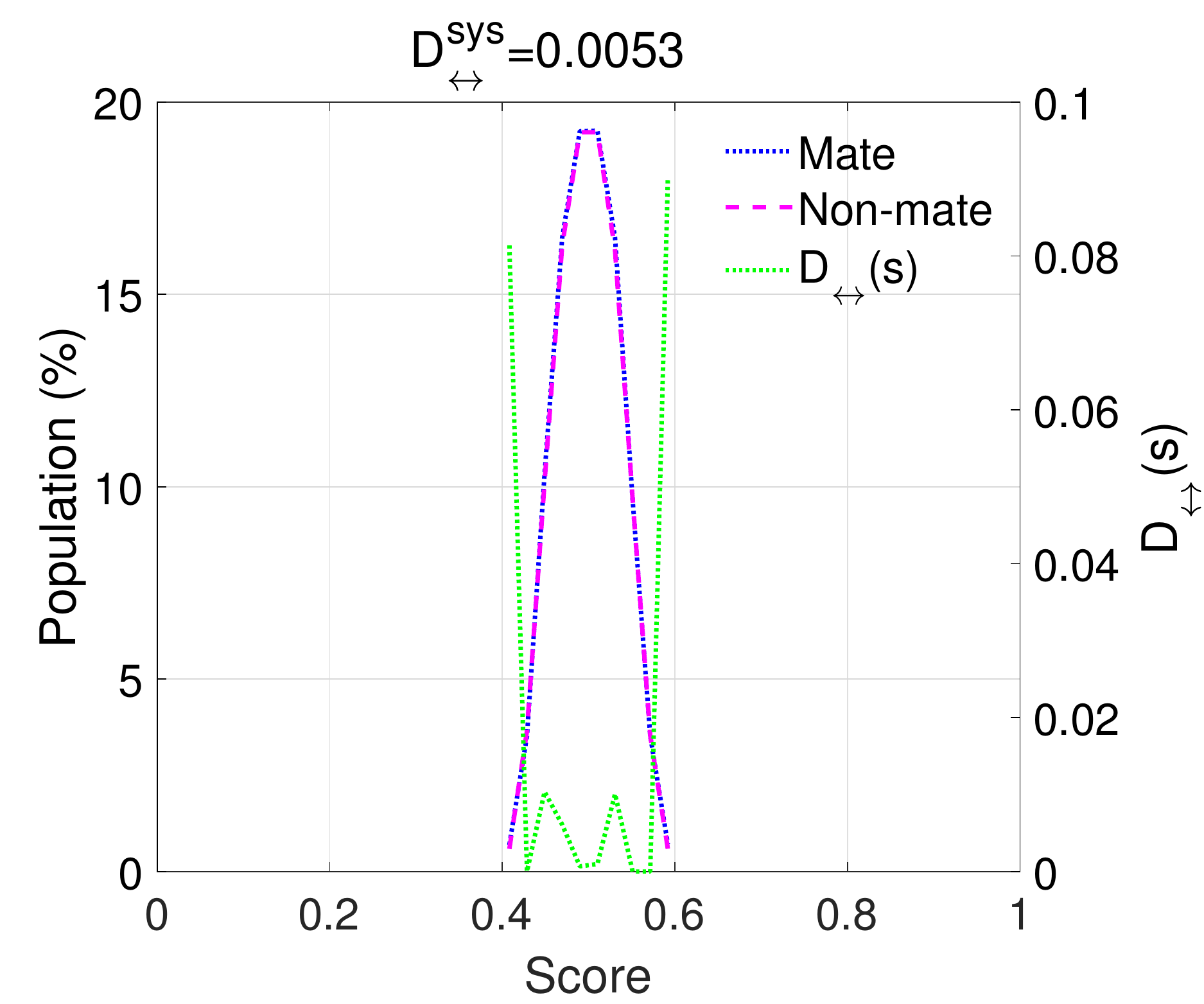}
		\caption{WhuGAIT dataset}
		\label{fig_unlink_whu}
	\end{subfigure}%
	\caption{The unlinkability score measured on the reliable string $\omega$.}
	\label{fig_unlink_score}
\end{figure*}

%
%
%
%
With each user in the testing sets, we randomly generated $100$ projection matrices, and used them to generate $100$ reliable strings. 
From the generated strings, we calculated the mated and non-mated scores, as well as $\text{D}_{\leftrightarrow}{(s)}$ and $\text{D}_{\leftrightarrow}^{sys}$.
Figures \ref{fig_unlink_ana} display the unlinkability scores measured on OU-ISIR and whuGAIT datasets. 
%
%
%
%
By the small global measure scores (\textit{i.e.}, $0.0028$ and $0.0053$ for OU-ISIR and whuGAIT, respectively), the unlinkability of our model is confirmed.


\subsubsection{Security Analysis}\label{ssec_secu_analyze}

We evaluate our model against typical attacks on existing BCS schemes.

\textit{Stolen Key Inversion Attack} \cite{scheirer2007cracking}: 
This attack bases on an assumption that the generated key has been compromised (which commonly occurs in practice). 
Then, the revealed key can be leveraged along with the helper data to learn a part (or all) of the biometric template.
Then, all other BCS models of the target user will be permanently compromised.
A BCS solution is vulnerable to this attack if it can not fulfill the irreversibility requirement.
As analyzed in \ref{sssec_security_improve}, in the IECO scheme, the attacker could not gain further advantage on reverting the original biometric data by observing the compromised key.
Thus, our model is immune to the key inversion attack.

\textit{ Record Multiplicity Attacks (RMA)}:
These attacks exploit the correlation between multiple instances of helper data/key extracted from a same person (\textit{e.g.}, FES \cite{simoens2009privacy}, FVS \cite{scheirer2007cracking}, cancellable biometric \cite{li2014attacks}).

In FES \cite{dodis2008fuzzy}, the XOR result of the biometric template and a codeword is stored as helper data. 
Then, given two helper data instances, the attacker can identify with high probability whether they are derived from a same user or not, by checking their XOR result \cite{simoens2009privacy}. 
This relied on a property of ECC which features that XOR-ing two valid codewords will result to a valid codeword.
Thus, the XOR-ing of two helper data instances from a same user will be close to a valid codeword with high probability.
It is clear that such attack is infeasible in our model, as we do not store the XOR result of the codeword and gait data.

In FVS, by matching two vaults (\textit{i.e.,} helper data) derived from the same user, the biometric features could be identified \cite{scheirer2007cracking}. 
This allows reconstructing the biometric data, and reveal the sealed key with high certainty. 
In our model, the biometric templates are covered by RP to provide unlinkability between different enrollments.
By matching two helper data instances, the attacker could not identify which locked points are from biometric features.
%
%
Thus, the strategy of RMA used in FVS does not work on our model.

RMA also can be used to compromise the scheme that relies on RP to fulfill the irreversibility \cite{li2014attacks}.
In such model, the irreversibility was addressed by a many-to-one linear system, in which, there are infinite solutions for a given pair of projected data and RP matrix.
However, when the attacker can obtain multiple instances of transformed template and projection matrix, the original template could be identified easily.
In our model, the transformed template is not stored, but used for key generation.
From the key, it is impractical to reconstruct the transformed template.
Without the transformed template, the attacker could not revert to the original biometric data.

\textit{Statistical Attack}: 
This attack targets the BCS models in which the extracted binary strings have low inter-class discrimination high intra-class variation \cite{rathgeb2011statistical}.
Due to the high intra-class variation, two-layer ECC is usually adopted to increase the error correcting capability \cite{hao2006combining}.
In such model, the biometric binary string is equally divided into several chunks.
The first ECC layer handles the error bit(s) in each chunk separately, then, the second layer corrects the fault chunks by checking the consistency over all the chunks.
%
Given the helper data, the attacker can try with a small biometric dataset. 
For each chunk, the histograms of output codewords is determined.
The codeword corresponding to the highest bin of the histogram is then selected as the guessed codeword for the corresponding chunk \cite{rathgeb2011statistical}.
In our model, due to the effectiveness of the deep model, the extracted binary string $\omega$ features low intra-class variation and high inter-class separation (see \ref{fig_hamd_eico}).
This allows the adoption of a long key and low-error-toleration BCS scheme (\textit{i.e.,} we use only one ECC layer).
Thus, our model is secure against the statistical attack \cite{stoianov2009security}.
\begin{table*}[t]
	\centering
	\caption{A comparison on different factors between our model and existing gait cryptosystems, where R, I, and U mean Revocability, Irreversibility, and Unlinkability, respectively.}
	\label{tab_comparison}
	\def\arraystretch{1.1}
	
	\begin{tabular}{ccccccc} \hlineB{3}
		\multirow{2}{*}{\textbf{\textit{Study}}} & \textbf{\textit{Data}}&\multirow{2}{*}{\textbf{\textit{Key size}}} & \multirow{2}{*}{\textbf{\textit{Performance}}} & \multirow{2}{*}{\textbf{\textit{R}}} &\multirow{2}{*}{\textbf{\textit{I}}}& \multirow{2}{*}{\textbf{\textit{U}}} \\
		&\textbf{\textit{Length}}& &&&& \\
		\hlineB{2}
		Hoang \textit{et al.} & \multirow{2}{*}{$16$ cycles} & \multirow{2}{*}{$139$ bits}&FAR: $0\%$  &\multirow{2}{*}{$-$}&\multirow{2}{*}{$-$}&\multirow{2}{*}{$-$}\\ 
		2015 \cite{hoang2015gait} &&&FRR: $16.18\%$&&&\\ \hline
		Tran \textit{et al.} &  \multirow{2}{*}{$26$ cycles}&  \multirow{2}{*}{$148$ bits}&FAR: $6\times 10^{-5}\%$  &\multirow{2}{*}{$-$}&\multirow{2}{*}{$-$}&\multirow{2}{*}{$-$}\\ 
		2017 \cite{tran2017improving} &&&FRR: $9.2\%$&&&\\ \hline
		%
		%
		\multirow{4}{*}{This study} & 	\multirow{4}{*}{$5$ cycles}&\multirow{2}{*}{$139$ bits} &  FAR: $0\%$ &\multirow{4}{*}{Yes}&\multirow{4}{*}{Yes}&\multirow{4}{*}{Yes}\\
		&&&FRR: $5.56\%$&&&\\ 
		&& 	\multirow{2}{*}{$147$ bits}   & FAR: $0\%$ &&&\\
		&&&FRR: $7.64\%$&&&\\ 
		
		\hlineB{3}
	\end{tabular}
\end{table*}
\subsubsection{Comparison with Existing Works}

To provide a meaningful comparison to existing gait cryptosystems \cite{hoang2015gait,tran2017improving}, we additionally experimented our model on CNU dataset \cite{hoang2015gait}, which has been used to evaluate their models.
However, the gyroscope signal is not available on this dataset. 
Thus, we modified the network to accept the gait segment of 3 channels as the input (instead of 6 channels as in \ref{sssec_netarch}).
Table \ref{tab_comparison} provides a comparison between our model and existing gait cryptosystems.
It could be seen that, our model not only achieved higher performance, but also required smaller data segment for key reproduction. 
In addition, unlike existing schemes, we allowed key revocation, and fulfill irreversibility and unlinkability. 

\section{Conclusions}\label{sec_conc}

This study presented a novel gait cryptosystem to generate from sensor gait data a key for user authentication, meanwhile, secure the gait pattern.
First, a deep network optimized by a new loss function was used to extract a representation template from the raw gait segment.
Then, RP and feature-wise binarization were subsequently used to extract a revocable binary string.
The string showed high stability when extracted from a same user, and great randomness between different users. 
Subsequently, an irreversible key was generated from the string, following the IECO scheme.
The evaluation on OU-ISIR and whuGAIT datasets showed that our model was more secure and efficient comparing to existing gait BCSs.
Moreover, our model was secure against existing biometric template protection attacks, and fulfilled the irreversibility and unlinkability requirements.

\section*{Acknowledgments}
The authors would like to thank Thang Hoang (Virginia Tech, USA) for helpful discussions and important references.
This work was supported in part by the Institute of Information \& Communications Technology Planning \& Evaluation (IITP) grant by the Korean Government, Ministry of Science and ICT (MSIT) under Grant 2020-0-00126, and in part by the Vietnam National University (VNU-HCM) under Grant NCM2019-18-01.

\bibliography{reference}

\end{document}